\pdfoutput=1
\documentclass[11pt]{article}
\usepackage[most]{tcolorbox}
\usepackage{microtype}
\usepackage{svg}
\usepackage{amsmath}
\usepackage{makecell} % for thicker lines
\usepackage{algorithm}
\usepackage[noend]{algpseudocode}
\usepackage{varwidth}
\usepackage{multirow}
\usepackage{tabularx}
\usepackage{subcaption}
\usepackage{amssymb}
\usepackage{amsmath}
\usepackage{amsthm}
\usepackage{amsfonts}
\usepackage{bbm}
\usepackage{bm}
\usepackage{booktabs}
\usepackage{graphicx}
\usepackage{multirow}
\usepackage{multicol}
\usepackage{enumitem}
\usepackage{mdwlist}
\usepackage{csquotes}
\usepackage{multirow}
\include{cv-commands}
\setlist{nosep}
\usepackage{soul}
\usepackage{mdframed}
\usepackage{style/acl2023}

\usepackage{times}
\usepackage{latexsym}
\usepackage{tabularx}

\definecolor{neverchanging}{RGB}{255, 227, 227}
\definecolor{slowchanging}{RGB}{255, 255, 179}
\definecolor{fastchanging}{RGB}{179, 255, 179}
\definecolor{falsepremise}{RGB}{179, 217, 255}
\usepackage[T1]{fontenc}
\usepackage[utf8]{inputenc}

\usepackage{microtype}

\usepackage{inconsolata}
\usepackage{adjustbox}

\title{\cfm{}: Cheap but Effective and Interpretable Answer Equivalence}

\author{
    Zongxia Li \hspace{3em} Ishani Mondal \hspace{3em} Huy Nghiem \\[0.25cm] % Adjust the 0.5cm to your desired spacing
    \textbf{Yijun Liang} \hspace{3em}\textbf{Jordan Boyd-Graber}  \\[0.25cm] % Same here
    University of Maryland, College Park\\[0.25cm] % And here
    \texttt{\{zli12321, imondal, nghiemh, yliang17, jbg\}@cs.umd.edu} 
}

\usepackage[a-1b]{pdfx}

\usepackage{framed}
\usepackage{mdwlist}
\usepackage{siunitx}
\usepackage{latexsym}
\usepackage{colortbl}
\usepackage{xcolor}
\usepackage{nicefrac}
\usepackage{booktabs}
\usepackage{fnpct}
\usepackage{amsfonts}
\usepackage[T1]{fontenc}
\usepackage{bold-extra}
\usepackage{amsmath}
\usepackage{amssymb}
\usepackage{bm}
\usepackage{graphicx}
\usepackage{mathtools}
\usepackage{microtype}
\usepackage{multirow}
\usepackage{multicol}
\usepackage{xpatch}
\usepackage{latexsym,comment}
\usepackage[normalem]{ulem}

\newcommand*{\missingreference}{{\Huge \colorbox{red}{?reference?}}}
\newcommand*{\missingcitation}{{\Huge \colorbox{red}{?citation?}}}

\makeatletter
\xpatchcmd{\@setref}{\bfseries}{\missingreference}{}{}
\def\@citex[#1]#2{\leavevmode
    \let\@citea\@empty
    \@cite{\@for\@citeb:=#2\do
        {\@citea\def\@citea{,\penalty\@m\ }%
            \edef\@citeb{\expandafter\@firstofone\@citeb\@empty}%
            \if@filesw\immediate\write\@auxout{\string\citation{\@citeb}}\fi
            \@ifundefined{b@\@citeb}{\hbox{\reset@font\missingcitation}%
                \G@refundefinedtrue
                \@latex@warning
                {Citation `\@citeb' on page \thepage \space undefined}}%
            {\@cite@ofmt{\csname b@\@citeb\endcsname}}}}{#1}}
\makeatother

\newcommand{\gem}[1]{\mbox{\textsc{gem}}}
\newcommand{\abr}[1]{\textsc{#1}}

\newcommand{\hidetext}[1]{}
\newcommand{\ignore}[1]{}
\newif\ifcomment
\commenttrue

\ifcomment
    \newcommand{\pinaforecomment}[3]{\colorbox{#1}{\parbox{.8\linewidth}{#2: #3}}}

    \newcommand{\prtodo}[1]{\pinaforecomment{lightblue}{pr}{#1}}
    \newcommand{\prtodoi}[1]{\pinaforecomment{lightblue}{pr}{#1}}
\else
    \newcommand{\pinaforecomment}[3]{}
    \newcommand{\prtodo}[1]{}
    \newcommand{\prtodoi}[1]{}
\fi

\newcommand{\smallurl}[1]{ \begin{tiny}\url{#1}\end{tiny}}

\definecolor{lightblue}{HTML}{3cc7ea}
\definecolor{CUgold}{HTML}{CFB87C}
\definecolor{grey}{rgb}{0.95,0.95,0.95}
\definecolor{ceil}{rgb}{0.57, 0.63, 0.81}
\definecolor{UMDred}{HTML}{ed1c24}
\definecolor{UMDyellow}{HTML}{ffc20e}

\definecolor{lightgray}{gray}{0.9}

\newcommand{\mm}[0]{\abr{llm}}
\newcommand{\muppet}[0]{large language models}
\newcommand{\bertscore}[0]{\abr{bert}Score}
\newcommand{\bert}{\abr{bert}}

\newcommand{\roberta}{\abr{r}o\abr{bert}a}
\newcommand{\jeopardy}{\textit{Jeopardy!\ }{}}
\newcommand{\cfm}{\textit{\abr{pedants}}}

\newcommand{\refer}{reference}
\newcommand{\can}{candidate}
\newcommand{\ac}{\abr{ac}}
\newcommand{\equivalence}{answer correctness}

\begin{document}
\maketitle

% \jbgcomment{This doesn't line up very well with PEDANT: Maybe something more
%   like Precise Evaluations of Diverse Answer Nominee Text for Skinflints (PEDANTS)  }

\begin{abstract}
Question answering (\abr{qa}) can only make progress if we know if an
answer is correct, but current answer correctness~(\ac{}) metrics struggle with verbose, free-form answers from \muppet{} (\mm{}s).
There are two challenges with current short-form \abr{qa} evaluations: a lack of diverse styles of evaluation data and an over-reliance on expensive and slow \mm{}s.
\mm{}-based scorers correlate better with humans, but this expensive task has only been tested on limited \abr{qa} datasets.
% , and compared with limited \abr{qa} metrics, but even when available, update of the model is limited because \mm{}
% are, well, large and often expensive.
%
We rectify these issues by providing
rubrics and datasets for evaluating machine \abr{qa} adopted from the
 Trivia community.
%
%We also introduce Precise ANswer-correctness Determination and Adjudication (\cfmatch{})---a small, efficient, deterministic \ac{} classifier (812 KB)---that more accurately evaluates answer correctness.
We also propose an efficient, and interpretable \abr{qa} evaluation that is more stable than an exact match and neural methods (\bertscore{}).\footnote{\url{https://github.com/zli12321/PEDANTS-LLM-Evaluation}}
\end{abstract}

\section{Introduction}
\label{section:intro}
% \zong{finetuned-bert on good data is better than gpt-4 for non-finetuned for classification~\cite{thalken2023modeling}}

% \zong{There are three general ways to improve \abr{qa} models' ability to better answer questions. The first way is to improve training data quality that better align with human judgments and need (cite less is more paper), but this can require a lot of labor intensive annotations and too broad and general to evaluate . The second way is to improve algorithm and model architecture and model size, which is computational expensive like LLM, and the hotness of LLM hinders people from developing other newer architectures from scratch, which is hard to get better than current transformer LLM that already has tons of human verification and training. The third way is to improve the evaluation metric, which is also one standard pipeline of \abr{qa}. Which the evaluation metric can be applied to any one of the first two. Even if we change model architectures, evaluation methods still can be applied.}

% \zong{Limitation: for subjective answers that have different answers for different demographics, we do not put them as one category and it deserves another new paper.}

% \abr{qa} is a key task in natural language processing; indeed, many language
% model interactions can be considered \abr{qa} tasks~\cite{sanh2022multitask}.
%
% There are two common ways to improve \abr{qa} models' ability to better answer
% questions: use new and better data or build new, better, and (usually)
% bigger models.
% %
% We do neither.
%
% Instead, we improve evaluation.
\abr{qa} evaluation is a necessary pipeline to train and optimize \abr{qa} models~\cite{sanh2022multitask}. 
%
% A skeptical reader might ask: won't a better model always have a higher score
% even if the evaluation is better\dots who cares?
%
% \jbgcomment{cite Vorhees}
From a model selection perspective, Section~\ref{subsec:pairwise ranking} shows that a weak evaluation can lead to incorrect conclusions about the ranking of models.
% ~\cite{voorhees-tice-2000-trec}
%
From a training perspective, having the right objective to train a model is an essential step to guide the model to the right direction~\citet{answerExpansion} shows that keeping models, data, and \emph{test} evaluation fixed, if we improve automatic evaluation during training time, the trained model improves on test set accuracy than using a more rigorous training evaluation-- Exact Match (\abr{em}).

\begin{figure}[t]
    \centering
    \hspace*{-0.2cm} % Shift the figure to the left by 1cm
    \includegraphics[width=0.8\linewidth]{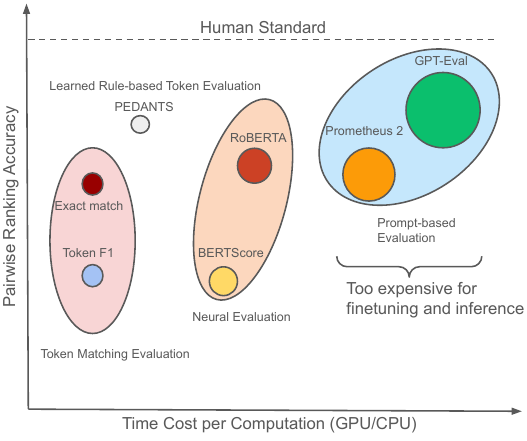}
    \caption{Different evaluation methods have different requirements of computation resources on short-form and factoid \abr{qa} datasets. Their pairwise ranking accuracies are based on our annotated data in Section~\ref{subsec:pairwise ranking}.}
    \label{fig:pairwise_acc}
\end{figure}

% you can still
% improve test evaluation by improving the \emph{training}
% evaluation~\cite{answerExpansion}: backpropagation relies on reliably
% identifying the examples your method got wrong.
% %
% If you're wrong about what you're wrong about, the trained model will be
% worse.

However, one of the reasons \abr{em} is popular
is because it is efficient and interpretable compared to other evaluations~\cite{kamalloo-etal-2023-evaluating}.
\emph{Twelve out of fourteen} papers (2022-2024) in Table~\ref{tab:qa_papers} involving \abr{qa} model training use either \abr{em} or $F_1$ as evaluators while one paper uses \abr{gpt-4}.
Thus, we focus on improving the efficiency, stability, and interpretability of short-form \abr{qa} evaluation: given a set of gold
answers, is the output of a system correct compared to the gold answers?
The standard \ac{} evaluations of \abr{qa} are \abr{em}---is the
answer string equal to a \refer{}---and token $F_{1}$ score---what
proportion of tokens match.
%
% Other measures are \abr{rouge} score~\cite{lin-2004-rouge}, etc, borrowed from the text translation and summarization literature.  
Neural evaluations such as \bert{}Score~\cite{Zhang2019BERTScoreET} and 
\bert{} Matching \cite[\abr{BEM}]{bulian2022tomayto} calculate pretrained contextual
embedding similarity scores between two string inputs.
\mm{}s as a judge such as \abr{gpt-4}, Claude~\cite{verga2024replacingjudgesjuriesevaluating}, or Prometheus~\cite{kim2024prometheus} use \mm{}s' reasoning ability and internal knowledge to judge \ac{}.

Standard (\abr{em}) and neural evaluations (\abr{bem}) are unstable on different styles of \abr{qa} datasets, where \abr{em} is only good for evaluating factoid answers and \abr{bem} is mostly good at answers that require reasoning.
Section~\ref{sec:human_evaluation} shows that only black-box \mm{} evaluators (GPT-4) are strong short-form \ac{} judges but they require much more cost and runtime than standard evaluations; open-sourced \mm{}s struggle at judging incorrect answers, generating many false positives.

% \jbgcomment{perhaps instead
%   of just citing original paper, it would be better to cite that *and*
%   where it's applied to QA?}
% While these popular \abr{qa} evaluation methods work okay on the most common \abr{qa} evaluations, Section~\ref{sec:human_evaluation} argues that they stumble on the different styles of test sets because they lack the depth of understanding of the contexts of the
% questions and answers that humans use when adjudicating answers.

% Section~\ref{sec:limitations} argues 
% that they stumble on the long tail of \abr{qa} examples because they
% lack the depth of semantic understanding of the contexts of the
% questions and answers that humans use when adjudicating answers.

% Fortunately, we are not starting from scratch! 
%
To come up with a consistent and structured short-form and factoid \abr{qa} evaluation, we revise existing rules from the Trivia community, integrating standardized \ac{} rubrics from \abr{naqt}~\cite{naqt_correctness} and the efficient \abr{qa} competitions~\cite{pmlr-v133-min21a}. 
The rules help us define machine \ac{}, especially for the challenging long tail scenarios (Section~\ref{section: classifier_training}). 
We also incorporate difficult \abr{qa} examples from the \textit{Jeopardy!\ }community to evaluate current \abr{qa} evaluation metrics (Appendix~\ref{sec:appendix_Jeopardy}).
We then use the our \abr{ac} framework to prompt GPT-4~\cite{bubeck2023sparks} and distil \abr{ac} knowledge to train a two-level classifier that goes beyond \abr{em} and provides a more fine-grained, interpretable and efficient \ac{} evaluation.
This broader view of \abr{qa} evaluation is necessary because modern \mm{} generate more verbose answers~\cite{qa_survey}, making \abr{em} less effective. 
Out of the examples we analyzed for \abr{em}, $90\%$ of answers have correct \can{} answers missing from \refer{}
answer sets, which humans consider correct but \abr{em} does not (Section~\ref{sec:error_analysis}). 

Our contributions to automatic \abr{ac} evaluation are:
\begin{enumerate}
    \item We adopt \abr{ac} rules from the Trivia community to build \abr{qa} evaluation framework.
    \item We propose \cfm{}: \textit{Precise Answer Normalizations}. We distil GPT-4 and improve the efficiency and effectiveness of short-form and factoid \abr{qa} evaluation.
    \item We evaluate popular \abr{qa} datasets and show that only black-box models are good evaluators across eight selected datasets with varying question styles. \cfm{} is more stable and effective than  \abr{em} and \abr{bem}. 
    \item We analyze hard examples that existing evaluations---token-based evaluations, neural evalations, prompt-based evaluations, \cfm{}---struggle with.
\end{enumerate}

\label{introduction}

% \input{sections/20-error-analysis}
% \label{baseline}

\section{Learning Evaluation From the Trivia Community}
% \jbgcomment{I worry that you have too much ``meat'' of the paper in
%   Table 1.  and the previous section might be seen as a ``related''
%   work.  So I think I'd put the clear contribution of the revised
%   correctness guidelines here and talk a little more about the process
%   of how you adopted them.  I'd also include more examples in the
%   appendix.}
 
%
% We then incorporate these insights into a \ac{} training set to induce new
% \ac{} classifiers.
Using \mm{}s to evaluate \abr{qa} is increasingly popular~\cite{chang2023surveyevaluationlargelanguage}.
However, existing \mm{} evaluations have inconsistent definitions, rubrics, and prompts to evaluate specific benchmark datasets~\cite{chiang2023closerlookautomaticevaluation,vu2024foundationalautoraterstaminglarge}.
Since \mm{}s are sensitive to change of prompts/definitions, comparing results among various \mm{} evaluation papers becomes hard. 
On the other hand, although \abr{em} is still the most popular standard evaluation, its lack of formal correctness definitions makes it unstable and less effective on more verbose \abr{qa} models (Section~\ref{sec:error_analysis}). The imperfections of current evaluations urges for starting rubrics and frameworks to evaluate models more fairly.

Luckily, we are not starting from scratch!
Because many of the \abr{qa} data and evaluations---from Watson on \textit{Jeopardy!}
to TriviaQA---come from the trivia community, \citet{rodriguez-boyd-graber-2021-evaluation} argues that researchers should not just take their data
but also examine their \emph{rules} to bridge the gap between \abr{nlp} and human community needs.
Moreover, the Trivia community reflects decades of evaluating \ac{} in high-stakes human competitions that need comprehensive and uniform
rules to ensure fairness.
Because the subject areas are meant to cover the standard school knowledge, the curriculum breadth reflects many of the evaluation challenges for \mm{}s.
While they are not perfect, they are more comprehensive than any existing
automatic \abr{qa} evaluation framework~\cite{bulian2022tomayto}.
We propose \ac{} rules from gold standard human \abr{qa}
evaluations: \abr{naqt}~\cite{naqt_correctness},
\textit{Jeopardy!}~\cite{carberry_jeopardy_casebook}, and
EfficientQA~\cite{pmlr-v133-min21a}.

% \jbgcomment{One thing that reviewers seemed to have doubts about why we should adopt these rules.  It might be good to do things like:

% Showing respect for community where we take the data from (and if we take their data we should follow their rules)
% https://www.sciencedirect.com/science/article/pii/S2666389922000228

% These are high-stakes competitions, fostering comprhensive and uniform rules

% The subject areas are meant to cover human knowledge, offering breadth

% While they are probably not perfect, they are better articulated than any existing AE framework
% }

\subsection{\ac{} Framework for Evaluation}
\label{sec: corectness_framework}

% Although we sketched some issues with existing \ac{} in the previous section,
% Enumerating all deficiencies to automate evaluation is intractable.
% %
% Fortunately, the Trivia community already has a set of
% well-defined norms to test human \abr{qa} ability.
%
\abr{naqt} is a thirty-year-old purveyor of \abr{qa} competitions spanning
middle schools to ``open'' competitions that anyone can join.
Because of the breadth of participants and judges (often drawn from parent
volunteers of participants), the organization has developed easy-to-follow
rules that make answer adjudication consistent (otherwise, participants would
complain about unfairness).
However, \abr{naqt} and \textit{Jeopardy!} formats are for
\emph{human} players, we cannot unquestioningly adopt the rules for machine
\abr{qa}.
For example, some rules cover in-person delivery of answers: mispronouncing
\underline{koan} as \underline{cone}, pauses between parts, etc.
These and other rules are irrelevant to text-based \abr{qa}.

%
% Likewise, the American game show \textit{Jeopardy!}---previously the setting of \abr{ibm}'s \textit{tour-de-force} \abr{qa}.

% \jbgcomment{I think I disagree with some of this.  Sure, pronunciation isn't applicable, but foreign language names I think is still certainly relevent, as is name order.  ``Li Zongxia'' is acceptable, but ''Graber Boyd Jordan'' is not.  So I think I'd just focus on pronunciation.

% \paragraph{Experts also play a role in machine \abr{qa} paradigm}
%

%

% \jbgcomment{I think rather than giving one concrete example here, I'd list all of the aspects of a correct answer here and forward point to the appendix where you can go into more detail.  I think we can also format the appendix guidelines more nicely, as right now it's the same as we used for a gpt prompt}

On the other hand, the rubrics nonetheless provide rules we can adopt.
For example, the \textit{specificity} rule is
both present in the \abr{naqt} and \textit{Jeopardy!} rules, where the
responses should be specific enough under the context of a
question---\textit{Question: Where is the Eiffel Tower located?}---where the
answer \textit{\uline{Europe}} is incorrect if the given \refer{} answer is
\textit{\uline{France}}, but is acceptable if \textit{\uline{Europe}} were the
intended answer.
% \jbgcomment{it would be good to have a typographical convention that
%   distinguishes questions from answers.  Perhaps make a macro; I like
%   underline for answers (perhaps inspired from NAQT)}

\begin{table}[t] % This specifies that the table should appear at the top of the page
\centering
\tiny
\renewcommand{\arraystretch}{1.5} % Increase row spacing
\begin{tabular}{p{2cm}p{4.8cm}} % Adjust the column widths to fit one column of the page
\hline
\textbf{Rule} & \textbf{Description} \\ \hline
$R_1$: Entity-aliasing & Widely recognized aliases, pseudonyms that are commonly associated with referred answer entities are acceptable. \\ 
$R_2$: Numerical information & Exact dates, years, numerical values are required unless the question specifically asks for approximations. \\ 
$R_3$: Less details & The answer provides less detail but should include the essential and correct information required by the question (specificity level 1). \\
$R_4$: More details & The answer contains additional information that does not contradict the question or initial answer (specificity level 2). \\
$R_5$: Semantic equivalence & A high degree of word overlap does not establish equivalence. The answer must be contextually and semantically accurate. \\
$R_6$: Irrelevant information & Any irrelevant or inaccurate description related to the question should be considered incorrect. \\
$R_7$: Other possible answers & The response is correct but not in the initially provided list. \\
\hline
\end{tabular}
\caption{The correctness of a \can{} answer can be traced and categorized to
  one or more of the above rules. We adopt the rules from \abr{naqt} modified
  after analysis of an annotated dataset with human correctness judgments
  between \refer{} answers and machine generated answers. The acceptability of
  \abr{qa} model answers are based on the rubrics. Rules with
  examples are in Table~\ref{tab:more_ae_examples} (Appendix).}

% \ishanicomment{In order to justify the adoption of these rules, we can also provide examples and a very short description of the rules, as was done by the Tomahto, Tomato paper? Something like your Figure 1 in Appendix}
\label{tab:guidelines}
\end{table}

\paragraph{The revising process} 
To find the rules that apply to text-based \ac{}, 
% we examine 1,688 examples from
% \abr{ropes} dataset and select 600 examples in the official \ac{} test set
% that have misalignment between human judgments and \abr{em}. 
We manually annotate 200 examples from the \ac{} test~\cite{bulian2022tomayto} set where \abr{bem} and human
judgments misalign.
We then find the rules from the trivia rubrics that explain the
disagreement and revise them to remove human-specific desiderata
(Table~\ref{tab:guidelines} with \abr{qa} examples in
Appendix~\ref{section:more_ae_examples}).

% and add machine \ac{} considerations (Table~\ref{tab:guidelines} with additional examples in Appendix~\ref{section:more_ae_examples}).

% \jbgcomment{don't have space after colon in labels}

% We enforce the professional \abr{naqt} \ac{} rules and guidelines and incorporate them into current machine \abr{qa} \ac{} paradigm. Then we provide an in-depth and comprehensive machine \abr{qa} correctness rules to access \ac{}. In addition, we not only delve into misalignment of \ac{} examples between human and selected automated evaluation methods, we also selectively analyze the aligned examples that all evaluation methods get right to ensure the generality and comprehensiveness of our \ac{} guidelines. Specifically, we analyze 1,688 examples from \abr{ropes} dataset and select 600 examples in the official \ac{} test set that have misalignment between human judgments and the other two evaluation methods. We also pick 200 additional examples from the \ac{} test set where both \abr{bem} and human judgments align. We present a comprehensive category of our modified correctness rules that can fit into the current machine \abr{qa} paradigm in Table ~\ref{tab: guidelines} and discuss and provide examples of acceptability according to these rules. 

\label{pipeline}

\section{Efficient \abr{qa} Evaluation}
\label{section: classifier_training}
% \ishanicomment{I think we can also cite some existing work which focused not only on performance but also these other factors in order to justify.}
% \zongxiacomment{Put pseudo code in the appendix}
Evaluation is not a task like \abr{qa} itself where the size of the model is a secondary consideration; an effective method (i.e., one that could be
adopted) needs to be accurate, fast, and low-cost~\cite{rieger2020irof}.
This section's goal is to match the effectiveness of the \mm{} evaluation
while minimizing computation, latency, and disk space.
\abr{em} and token $F_1$ evaluations are fast and require no storage but are superficial.
\mm{} evaluations are trainable but require substantial disk space, cost, computational time, and latency.
In aggregate, they might be too expensive for universal adoption in standard \abr{qa} training and evaluation pipelines.

By revising \ac{} for text-based \abr{qa}, we can construct an efficient
automatic evaluation that not only aligns with our \ac{} decision-making
process but also goes beyond token-level matching that follows the human mental judgment process.

\subsection{\cfm{} Details}
% \ishanicomment{Figure 1 and notations look clear now. What if we do something like, creating a process diagram, where we first show these two examples with your ($q$, $a$, $\tilde{a}$), as input, determine $T$ and $R_n$ after that, show how do we construct representative training data (maybe use rulebooks as your guide, give an arrow pointing GPT-4 to generate more data), and finally do the feature engineering. This is just to make the input and output for each different step very clear. Name these steps in your flowchart, and in the body of the text, just refer back to the same names in the figure so that reviewers can look up the input and output in a very short amount of time?}
This section details how we train the \cfm{} pipeline.
This is a two-step classifier that first predicts the 
type $T$ of the question and which of the rules $R$ from Table~\ref{tab:guidelines} are applicable.
It then creates a final decision of whether given a question~\(q\), a reference answer~\(a\), and a candidate answer~\(\tilde{a}\), \cfm{} classifier decides whether a candidate answer is correct or not.
We first define what question types and \abr{ac} rules, describe how we train those preliminary classifiers, and then describe the training of the final classifier.

% \ishanicomment{Connect back to your Joe Biden example in Figure 1.}

\paragraph{Question Type}
Let \( T \) denotes the type of a question, categorized by \{\textit{who}, \textit{why}, \textit{how}, \textit{what}, \textit{when}, \textit{where}, \textit{which}\}. 
Specifically, the type of a question~\( q \) primarily depends on the content of~\( q \) itself, but it may also depend on the context of reference answer~\( a \). 
Therefore, we define the type \( T \) of the pair~\((q, a)\) as one of the type categorizations.

\paragraph{Applicable Rules}
Given pair ($q$, $a$, $\tilde{a}$), $R_n$ is the rule from Table~\ref{tab:guidelines} used to judge the correctness of $\tilde{a}$.
For example, if the question is asking about \textit{\underline{Joseph Robinette Biden}}, then at a minimum the answer must contain \textit{\underline{Biden}} (the family name), but \textit{Robinette} would not be enough.

% \paragraph{Token $F_1$ Score} Token $F_1$, precision and recall measures the word overlapping similarity between ($a$, $\tilde{a}$), which we define as $TF(a, \tilde{a})$.

% \begin{figure}[!t]
%     \makebox[\linewidth][c]{%
%         % \hspace{-0.5cm}  % Adjust X to the desired value
%         % \includegraphics[scale=0.28]{figures/new_pipeline.pdf}
%         \includegraphics[scale=0.6]{figures/EMNLP_demo.pdf}
%     }
%     % \caption{Using text input and raw token scores as features can help \cfm{} learn an optimal combination of $F_1$, precision, and recall score thresholds to judge different types of answers based on implicit learned rules.}
%     % \jbgcomment{If this could be compressed into one tall figure, it would save some whitespace}
%     \caption{Under different types of questions and different evaluation rules, even thought the token $F_1$ scores are the same for the above two examples, the judgment decision is different.}

%     \label{fig:pipeline}
% \end{figure}

\begin{table}[t]
\centering
\tiny
\footnotesize
\begin{tabular}{>{\raggedright\arraybackslash}p{6cm}}
\hline
$\bullet$ $q_1$: Who is the president of the US in 2023? \\
$\bullet$ $a_1$: Joe Biden \\
$\bullet$ $\tilde{a}_1$: Joseph Biden \\
$\bullet$ $q_1$ is asking about \textit{who}, having a question type $T_{who}$.\\
$\bullet$ $(q_1, a_1, \tilde{a}_1)$ is classified as Rule $R_1$-commonly known entity aliases are equal\\
$\bullet$ Token $F_1$ : 0.5, precision 0.5, recall 0.5. \\
$\bullet$ Under the constraint of $R_1$ and $T_{who}$ $\rightarrow$ \textbf{correct} \\ \hline
$\bullet$ $q_2$: When did Joe Biden become the president of the US? \\
$\bullet$ $a_2$: Jan 20, 2021 \\
$\bullet$ $\tilde{a}_2$: 2021 \\
$\bullet$ $q_1$ is asking about \textit{when}, having a question type $T_{when}$.\\
$\bullet$ $(q_2, a_2, \tilde{a}_2)$ is about numerical dates and years--$R_2$\\
$\bullet$ Token $F_2$: 0.5, precision 0.33, recall 1.0. \\
$\bullet$ Dates and years should have an exact match $\rightarrow$ \textbf{incorrect} \\ \hline
\end{tabular}
\caption{Despite identical token \( F_1 \) scores for the two \abr{qa} examples, the judgment decisions vary due to different question types and evaluation rules.}
\label{fig:pipeline}
\end{table}

Both of these are necessary prerequisites before we can make a final \ac{} decision.
This is because \emph{which} and \emph{how many} tokens must overlap between reference and candidate answers depends on the type of questions and applicable rules.

\paragraph{Mirroring human judgments}
Table~\ref{fig:pipeline} shows that the necessary token overlap for determining \ac{} varies with the question type \(T\) and rule \(R_n\), mirroring the variability in human judgment.
In this context, we refer to \(T\) and \(R_n\) as our question type features and rule features.
An automated answer \ac{} judgment needs to balance flexibility with specific constraints to ensure accurate judgments on various edge cases:
\begin{enumerate}
    \item Person names and entities may match despite low token overlapping scores, as variations in aliases (e.g. \textit{Joe} vs. \textit{Joseph}) can obscure lexical likeness.
    \item If the answer includes extra details but the core information is correct, we need a lower token matching threshold and high recall to ensure the essential information is answered (e.g. \textit{France} vs. \textit{France, Europe}).
    \item For date and numerical answers, stricter token matching is required, while type features ($T$) becomes important for identifying numerical answers, etc.
\end{enumerate}

In the evaluation of ($q_1, a_1, \tilde{a_1}$), the feature $R_1$ and $T_{who}$ allow for correctness judgments even when the candidate answer $\tilde{a_1}$ and the reference answer $a_1$ do not exactly match or share high token $F_1$, precision, or recall scores. 
Conversely, exactness in dates is required for ($q_2, a_2, \tilde{a_2}$) categorized under feature $T_{when}$ and $R_2$. 
Hence, even though the token $F_1$ match between $a_2$ and $\tilde{a_2}$ is similar to that of $a_1$ and $\tilde{a_1}$, the evaluation penalizes the lack of exactness in date answers (Table~\ref{fig:pipeline}). 
The combination of token $F_1$, precision, and recall thresholds is tailored to the specific question type and rule.
Therefore, features such as $T$, $R$, \textsc{tf-idf} encodings, and \emph{standard metric thresholds can be learned from data} that is representative of the \ac{} rules.

\paragraph{Training Data Preparation}
We manually write five to ten representative \abr{qa} examples with human
judgments as seed examples. 
% \ishanicomment{Connect back to the section where you mention this error analysis; it might look out-of-place.}
%
Then we prompt GPT-4 with the seed examples that contain type $T$ and rule $R_n$ and ask it to self-verify its generated judgment, $T$, and $R_n$ assignment~\cite{ren2023selfevaluation}.\footnote{Asking GPT-4 to self-verify its generation can improve the quality of its generations. Prompt Templates in Appendix Table~\ref{table:generation_prompt_template}.}
In the \ac{} training data, each example is paired with $(q, a, \tilde{a}, T, R_n)$, where $T$ and $R_n$ are categorical variables initially, but are treated as distribution of predictions when training the feature extraction classifiers.
%
% We manually annotate 350 generated examples and keep 3,989 \ac{} examples.
%
We then prompt GPT-4 to assign $T$ and $R_n$ to all examples in \citet{bulian2022tomayto}'s \ac{} training set (9,090 examples). 
We merged our generated data with \citet{bulian2022tomayto}'s examples as \cfm{}' training set (exact details and prompt templates in
Appendix~\ref{sec:data_augmentation}).

\paragraph{Training \cfm{}} 
% We first calculate the token $F_1$, precision, and recall for each $(a, a_n)$ in the training set, which are in 32-bit floats. 
% %
% We then use trained classifiers $F(R)$ and $F(T)$ to assign \ac{} rule and question type feature vectors to each training example. 
% %
% We use tf-idf to encode $(q, a, \tilde{a})$ the string \textit{[CLS] q [SEP] $a$ [SEP] $\tilde{a}$}, and concatenate $F(R)$ and $F(T)$'s out vectors, and token $F_1$, precision, recalls as inputs to train a logistic regression classifier (\ac{} judgment as outputs).\footnote{We concatenate all the features in the order \textit{[tf-idf encoding([CLS] q [SEP] $a$ [SEP] $\tilde{a}$), $F(R)$ output vector, $F(T)$ output vector, token $F_1$, precision, recall]} for each training examples as the classifier input. Every independent feature is a list of 32-bit floats or integers. Exact classifier hyperparameter training details are in Appendix~\ref{section: classifier_training}.}
% Now that we have the type and applicable rules, we can make the final decision of whether the answer candidate is correct.
%
We train two feature extractors, $F(R)$ and $F(T)$, to extract features for rule and question type features.
For each pair $(q, a, \tilde{a})$ in the training set, we lemmatize, remove punctuation, and calculate token $F_1$, precision, and recall. 
We encode $(q, a, \tilde{a})$ as \textit{[CLS] q [SEP] $a$ [SEP] $\tilde{a}$} using \textsc{tf-idf}, and concatenate the outputs from $F(R)$, $F(T)$, and the token scores to train a logistic regression classifier. 
Human annotations or GPT-4 judgments are used as \abr{ac} labels. 
For example, if $q$ is \textit{Who is the president of the US in 2023?}, with $a$ as \textit{Joe Biden} and $\tilde{a}$ as \textit{Joseph Biden}, $F(R)$ classifies the rule as $R_1$ (entity aliasing), and the question type as $T_{who}$, indicating a person.
Feature extractors are also logistic regression classifiers.
Details on classifier hyperparameters are provided in Appendix~\ref{section: classifier_training}.

In sum, \cfm{} is a pipeline classifier, with a feature extraction stage where \textsc{tf-idf} encoded ($q, a$) and $(q, a, \tilde{a})$ are processed to extract rule \(R\) and type \(T\) features. 
This is followed by a decision-making stage where $T$ and $R$ features, along with token \(F_1\), precision, and recall and \textsc{tf-idf} encodings, are input into a final logistic regression classifier to predict \ac{}. \cfm{}'s architecture simulates human judgment by integrating linguistic and rule-based features.

\label{methology}

\section{Evaluation Setup}
\label{sec:human_evaluation}
% While the previous section discussed stability and resources, we have not yet argued our method is \emph{better} or aligns with human judgements. 
% This section elaborates the process on how we generate a diverse style evaluation sets using six diverse language models ranging from 3B to 70B and black-box models. 

An effective evaluation of evaluation methods requires validations on diverse data and models.
% \jbgcomment{This section could be stronger if it connected to the big story.
%   How does this help us get to PEDANT?}
% \abr{em} is overly strict and inflexible. 
% %
% \bertscore{} and token $F_1$ require a pre-determined threshold prior to evaluation that introduces biases. 
% %
% % Fine-tuned neural models are good, but they can be affected by strong prior knowledge for short-form answers and require access to GPU resources. 
% %
% Black-box \mm{}s are nontransparent and high latency. 
%
We want to test limitations automatic evaluations on many datasets and \abr{qa} models to 1: verify their effectiveness on different styles of datasets, evaluating its strengths over other metrics and weaknesses regarding different the styles of answers; 2: evaluate their stabilization across answers generated by various \abr{qa} models (different sizes, different pretraining data, black-box and open-source models...) on the same dataset, which is important to provide a more fair comparison to \abr{qa} models on a benchmark; 3: ensure the test data is not restricted by the patterns of the training data to prevent overfitting on learned metrics.

% We need diverse answers from diverse models to prevent overfitting to the
% patterns of a specific model.
% %
% Likewise, we need diverse datasets to prevent overfitting to domains or
% \refer{} formats.
% %
% We selectively annotate examples and compare seven evaluation metrics against
% human judgments.
% %
% We emphasize the computation resources, costs, and time needed for each method
% to ensure the methods are usable not just for evaluation but also for model
% training.

% \jbgcomment{always spell out numbers $<10$.}

% \subsection{What Are Good Evaluation Sets?}
\paragraph{Dataset selection} 
We select three benchmark datasets that have factoid and short-form answers, four benchmark data that have answers in phrases or sentences (not summarization), and one with human generated answers.
The three factoid \abr{qa} datasets are sampled from \abr{nq-open}~\cite{kwiatkowski-etal-2019-natural},
\abr{NarrativeQA}~\cite{kocisky-etal-2018-narrativeqa}, and
HotpotQA~\cite{yang-etal-2018-hotpotqa}.
These three datasets are standard \abr{qa} benchmarks, used to train many
\abr{qa} models. 
The most popular metrics for evaluating the models on these datasets are \abr{em} and token $F_1$ score.
We use these datasets to verify that \cfm{} is at least as good as \abr{em}
and \bertscore{} on short-form \abr{qa} datasets.
% We select subsets of \abr{nq-open}~\cite{kwiatkowski-etal-2019-natural},
% \abr{Narrative \abr{qa}}~\cite{kocisky-etal-2018-narrativeqa}, and Hotpot
% \abr{qa}~\cite{yang-etal-2018-hotpotqa} questions as short-form test
% sets. These three datasets are classical \abr{qa} benchmarks that are used
% to train many \abr{qa} models, with exact match and $F_1$ score being the
% most popular evaluation for individual question and answers. We use these
% datasets to ensure \cfm{} works at least as good as \abr{em} and
% \bertscore{} on easy cases of \abr{qa} evaluation.
%

We also sample from \textit{Biomedical Machine Reading Comprehension} \cite[\abr{biomrc}]{stavropoulos2020biomrc}, \textit{Microsoft Machine Reading Comprehension} \cite[\abr{ms marco}]{bajaj2018msmarco}, \textit{Conversational Question Answering Challenge} \cite[\abr{coqa}]{reddy2019coqa}, and \textit{MOdeling Correctness with Human
Annotations} \cite[\abr{mocha}]{chen_2020} to test more challenging evaluation, where the answers are sentences that can confound \abr{em}.\footnote{We show failure modes of token matching for evaluation and that \cfm{} is more robust than neural methods like \bertscore{} for datasets with longer gold answers.}
%
% These datasets feature more verbose \refer{} answers, requiring a deeper
% lexical understanding of the relationships between words and sentences for
% effective evaluation.
% %
% The answer complexity poses more challenge to \abr{em} and token-based
% evaluation methods.
% %
% By using these datasets, we show failure modes of superficial token matching
% for evaluation and demonstrate that \cfm{} is more robust than neural methods
% like \bertscore{}.
%
% \jbgcomment{I worry that the the framing of this dataset lets people discount
%   it as a contribution.  Give a short description of it and say that full
%   documentation is in the appendix.  But don't say that the dataset ``is in
%   the appendix'', as that makes it seem like it isn't real.}

Furthermore, we collect a dataset from \jeopardy{}, featuring answers from real human players. 
This dataset challenges existing evaluation metrics such as \abr{em}, which by design scores 0 in Macro $F_1$ evaluations. 
Evaluating these answers correctly is hard and requires years of human experience.\footnote{\url{https://www.j-archive.com}. More details of these evaluation sets, including the number of
examples, example questions and answers are provided in
Table~\ref{tab:dataset} and Appendix~\ref{sec:appendix_Jeopardy}.}

%

%
% The eight test sets cover diverse style of questions and are not relevant to
% the training data, and covers the easy and hard cases of \abr{qa} evaluation.
%
% We concatenate the contexts with the question as inputs for all datasets.

% \jbgcomment{I think this hides the Jeopardy! dataset too much.  I think it's fine to put more of the details there, but I think the results on this dataset need to be in the main paper.  I think it's important to focus on how these are things that smart humans got wrong (both the answers and the AC evaluation).\zongxiacomment{This is a good idea. I would include some examples here and say that even GPT-4 cannot solve this by commensense reasoning.}}

\paragraph{Model Selection}
Different \abr{qa} models pretrained with different data generate different answers, and their generations can also
depend on model sizes.
We select a range of models to generate answers for selected datasets Flan-T5-xl (3B)~\cite{flan-t5}, LLaMA 2 (7B)~\cite{touvron2023llama},  \abr{gpt-3.5-turbo} (a black-box model),  Mistral 8x7b Instruct~\cite{jiang2024mixtral}, Yi-Chat 34B~\cite{ai2024yi}, and LLaMA 2
70B~\cite{touvron2023llama}.
Our models span 3B to 70B include both black-box and
open-sourced architectures, and incorporates models pretrained with different sources.

% Different \abr{qa} models fine-tuned with different training sets have diverse styles of answer generations. 
% %
% The generated answer qualities can also be highly dependent on model sizes. 
% %
% We select Flan-T5-xl (3B)~\cite{flan-t5}, LLaMA 2 (7B)~\cite{touvron2023llama}, \abr{gpt-3.5-turbo} (black-box model) as our \abr{qa} models to generate answers for \abr{nq-open}, Hotpot \abr{qa}, and \abr{narrative qa}.\footnote{\url{https://openai.com}} 
% %
% We select Mistral 8x7b instruct~\cite{jiang2024mixtral}, Yi-Chat 34B~\cite{ai2024yi}, and LLaMA 2 70B~\cite{touvron2023llama} to generate answers for \abr{biomrc}, \abr{ms marco}, and \abr{coqa}.
% %
% We add contexts if available. 
% %
% In addition, our selection of \abr{qa} models covers models from small sizes (3B) to large sizes (70B), from black-box models to open-sourced models, and includes models fine-tuned by different organizations to maximize answer generation styles.
% %
% We include open-source and black-box \abr{qa} models to ensure diversity of candidate answers. 

\paragraph{Evaluation metric selections} 
We choose evaluations from three mainstream methods to compare with \cfm{}: token evaluation, neural evaluation, prompt-based evaluation. Specifically, token evaluations are \abr{em} and token $F_1$, which compare the similarity of two strings; neural evaluations are \bertscore{}~\cite{Zhang2019BERTScoreET}, and \roberta{} matching~\cite{bulian2022tomayto}, 
% and \textit{Learned Evaluation metric for Reading Comprehension}~\cite{Chen_2020}, 
which use pretrained or finetuned transformer embeddings to measure the similarity between two strings; prompt-based evaluations are \abr{gpt-4} and Prometheus-2~\cite{kim2024prometheus}, which use finetuned generative models to generate an evaluation score.\footnote{Prompt templates for \abr{gpt-4} are in Appendix Table~\ref{table:eval_prompt_template}}
\cfm{} uses question and answer types to judge answer correctness with learned optimal string similarity matching.

% we use \abr{em}, token $F_1$, \bertscore{}, \abr{lerc} (threshold 4), 
% %
% \roberta{} and \cfm{} trained on 
% % original \ac{} training set-- \roberta{} and \cfm{}, 
% %
% our \ac{} training set, 
% % LLaMA 2 70B-instruct evaluation--LLaMA70B-Eval, 
% Prometheus 2~\cite{kim2024prometheus} (a fine-tuned 7B \mm{} evaluator), 
% and
% \abr{gpt-4} evaluation--
% \abr{gpt-4}-Eval.\footnote{\url{https://huggingface.co/upstage/Llama-2-70b-instruct}}
% Specifically, we prompt \abr{gpt-4} and Prometheus 2 with our \ac{} rules and
% demonstration examples to judge answer correctness (Prompt template in
% Table~\ref{table:eval_prompt_template}, Appendix).

% \section{Human Agreement}
% \label{sec:metric_comparison}

\subsection{Annotated Dataset Details}
Since \abr{mocha} is a benchmark evaluation dataset, it includes answers generated by GPT-2 and human rating annotations (Likert scale 1-5). 
We treat an answer as correct if it meets a threshold score of 4.

The \jeopardy{} data is a real-world \abr{qa} dataset with expert correctness judgements. 
Each example in the dataset has a question, a gold answer, a response by \textit{Jeopardy}{}! players, and expert judgment of the response. 
The dataset includes candidate answers that are challenging for experts to judge.
For example, given a question \textit{Your surgeon could choose to take a look inside you with this type of fiber optic instrument}, the gold answer is \textit{\uline{laparoscope}}, but \textit{\uline{endoscope}} was given and ruled incorrect during the show that later reverted to correct, verified by professionals in accordance with the \textit{Jeopardy}{}! answer acceptability rules. 
The \jeopardy{} dataset is meant to challenge both human judges and automatic evaluations to show the imperfections of automatic evaluations.
We collect 504 examples with examples in Appendix~\ref{sec:appendix_Jeopardy}.

\subsection{Annotating Generated Data}
Only \abr{mocha} and \jeopardy{}have available answers and judgements, so we use \abr{gpt-3.5-turbo}, Flan-T5 xl, and LLaMA 2 7B to generate answers for the three factoid and short-form datasets--\abr{nq-open}, HotpotQA, and \abr{nqa}; we also use Mistral 8x7b, Yi-Chat 34B, and LLaMA 2 70B to generate answers for challenging datasets--\abr{coqa}, \abr{biomrc}, and \abr{ms marco}.
We filter out examples where the reference and candidate answers do not have an exact match.
The total number of filtered examples is 42,090.
We then hire annotators on Prolific whose first language is English, with at least
a community college degree completed and above 99\% approval rate to judge the correctness of the filtered examples using our \ac{} rubrics in Table~\ref{tab:guidelines}. 
Out of 42,090 examples, 6,626 examples have annotations from two annotators, and the interannotator agreement of the 6,626 examples has Krippendorff's $\alpha=0.75$.
For the examples that have two annotations where annotators disagree, we have the authors annotate them to break tie the judgments.

\label{evaluation}

\begin{figure*}[t]
    \centering
    \includegraphics[width=\textwidth]{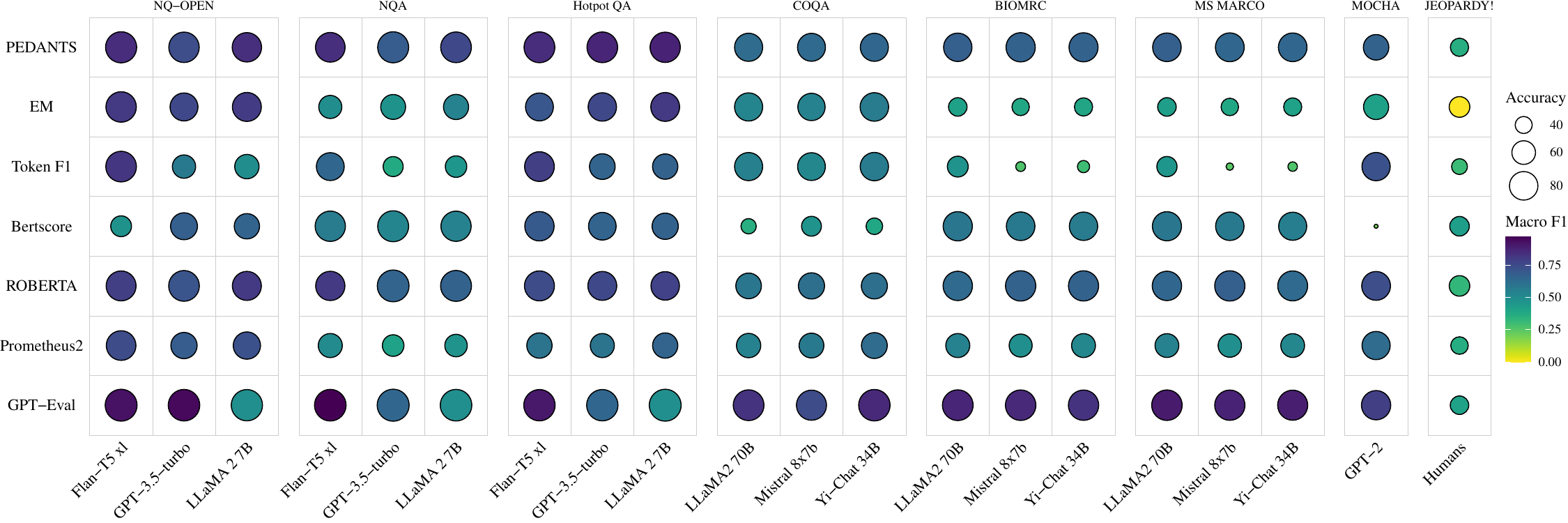}
    \caption{The size of the circles shows each metrics' human agreement accuracy and the color shows the Macro $F_1$ score . We put \cfm{} first for ease of visualization. \abr{em}, token $F_1$, and \bertscore{} have unstable human agreement on different \abr{qa} datasets by looking at horizontal circle size variations-- ranging from $25\%$ to $90\%$. \roberta{}, \cfm{}, and \abr{gpt4}-Eval are more robust and stable with varying \abr{qa} datasets. Although Prometheus 2 is fine-tuned for evaluation purposes, it fails on short-form \abr{qa}. \cfm{} is less costly than \abr{gpt-4}, Prometheus 2, \roberta{}, and \bertscore{} and has more stable human agreements across seven evaluation datasets than \abr{em}. Prometheus assigns scores from 1 to 5; we use 4 or higher as indicative of correctness.}
    \label{fig:metric_heatmap}
\end{figure*}

\section{Error Analysis}
\label{sec:error_analysis}
With available datasets with human judgments, this section analyzes the human agreement, time and monetary cost, and model ranking accuracy.
Then we dive into examples that are challenging to the standard evaluation methods (token matching and neural evaluation) but can be handled by \cfm{}.

\subsection{Human Agreement Rates}
\label{subsec: human_agreeement_rates}
For each dataset, we calculate each selected metric's accuracy and Macro $F_1$ score between automatic judgements and human judgments.
\abr{em} and token $F_1$ have high human
agreement on short \abr{qa} types---\abr{nq-open} and Hotpot \abr{qa}---but
falters for more challenging \abr{qa} datasets (Figure~\ref{fig:metric_heatmap}).
\bertscore{} can handle the more challenging datasets that have longer gold answers but falls behind token matching on short-\abr{qa}.
%
% \jbgcomment{Below sentence needs to be rewritten.  ``relative small marginal
%   agreement difference'' is hard to understand.  Write simply: ``about the
%   same'' is fine if that's what you're trying to say.}
\roberta{}, \cfm{}, and \abr{gpt-4} have similar and stable human agreement. 
\cfm{} is better than \roberta{} for short-form \abr{qa} datasets, but \roberta{} is better for long-form datasets that require a semantic understanding of word relationships.

Since Prometheus 2 is finetuned on evaluation data with long reference answers, it shows low human agreement on short-form \abr{qa} datasets by frequently misjudging simple cases such as the candidate \textit{\uline{No.}~versus} the gold answer \textit{\uline{No}} as incorrect.
However, Prometheus has higher human agreements than \abr{em} on more challenging long-form \abr{qa} datasets, which shows that prompt-based evaluations are not stable across different datasets.
Therefore, a more expensive method is not always better than a cheap method.
\subsection{Prompt-Based Evaluation Are Costly}
Token and neural evaluations are cheap but are inflexible.
Prompt-based evaluation like GPT-4 are the best if ignoring costs.
However, computation cost and time are important in real life practices, particularly if evaluation is a step during \abr{qa} model training.
Prompt-based evaluations are much more expensive than other evaluations regarding runtime and monetary costs (Table~\ref{tab:runtime}).
Open-source models like Prometheus require extensive GPU resources.
Since model training is already a GPU intensive and time-consuming task, an extra GPU for evaluation might not always be available or possible.
Although \abr{api}-based services like GPT-4 require no GPU, its generations often change with periodic updates~\cite{article}, leading to inconsistent evaluations.
In addition, GPT-4 API calls have high latency, requires network connection, and banned in fifteen countries as of 2024~\cite{Click4Assistance2023}, which is expensive and not accessible for every researcher to reproduce results.
In a realistic case, suppose a user who has 10,000 training data tries to train three models and pick out the best one, for a three epoch training, the total GPT-4 evaluation time is 21 hours and costs \emph{a thousant} dollars and approximately 7 days using Prometheus on an A6000 GPU.
On the other hand, one of the reason that \abr{em} or token $F_1$ are popular is that they are fast and easy to implement with minimum costs, but they are also less effective than prompt-based evaluations and \cfm{}.

\begin{table}[t]
\tiny
\centering
\begin{tabular}{cccc}
\hline
\rowcolor{gray!50}
Metric & Runtime/(min) & Disk Space & Cost  \\ \hline
\abr{em} & 0.05 & 0 & 0\\ 
Token $F_1$ & 0.05 & 0 & 0\\ 
% \cfm{} & 30 s & 7 MB & 0\\
\cfm{} & 1 & 5 MB & 0\\
\bertscore{} & 4.95 & 499 M & 0\\
% \roberta{} & 25 min & 499 M & 0\\
\roberta{} & 4.95 & 499 M & 0\\
Prometheus 2 & 1,123 & 12 GB & 0\\
\abr{gpt}-Eval & 140 & unknown & \$120\\
\hline
\end{tabular}
\caption{Though \abr{gpt-4} is the best evaluation model, it costs more money
  and time. The runtime is based on per 10,000 examples from 8 test
  datasets on 7 \abr{qa} model. \bertscore{}, \abr{lerc}, \roberta{}, and prometheus 2 evaluation methods
  are run on a single A6000 GPU device, and other methods are run with local CPU. The API cost for \abr{gpt-4}-Eval
  is about \$120.}
\label{tab:runtime}
\end{table}

\subsection{Pairwise Ranking Accuracy}
\label{subsec:pairwise ranking}
Model ranking is one of the most important use of evaluation. 
Human evaluations are expensive and time consuming, and they are usually used for big development cycles.
However, automatic evaluations are usually used during development cycles.
Thus, knowing which version of model is better on a task is essential to guide the model into the right development direction and prepare better training and evaluation data.

For each dataset with responses from $N$ models, we define pairwise ranking (\abr{pwr}) accuracy as the percentage of model pairs that an automatic metric have the same ranking as human rankings:
\begin{equation}
\frac{1}{\binom{N}{2}} \sum_{i=1}^{N-1} \sum_{j=i+1}^{N}[\left(\text{rank}_{\text{metric}}(i, j) = \text{rank}_{\text{human}}(i, j)\right)]
\label{eq:pairwise_accuracy}
\end{equation}
Where $(\text{rank}_{\text{metric}}(i, j) = \text{rank}_{\text{human}}(i, j)$ is 0 if automatic metric disagrees with human rankings, and 1 otherwise.

We calculate each metric's pairwise ranking accuracy against human rankings for datasets that have responses from multiple models.
GPT-Eval and \cfm{} are more stable at ranking models correctly on simple and more challenging \abr{qa} datasets.
\cfm{} ranks all models correctly on four datasets (Figure~\ref{fig:pairwise_acc}).\footnote{We use the percentage of responses humans or a metrics judge as \textit{correct} on a dataset to assign a score (between 0 to 100\%) to a model, then rank models with descending scores.}
Surprisingly, while \bertscore{} show inconsistent rankings on more challenging datasets, it has higher human agreements than \abr{em} and Prometheus (Figure~\ref{fig:metric_heatmap}).
From answer-level comparison (human agreements) and model-level comparison (pairwise ranking) of evaluation metrics, GPT-Eval is the most robust and recommended evaluation, followed by \cfm{}.

\begin{figure}[t]
    \centering
    \hspace*{-0.2cm} % Shift the figure to the left by 1cm
    \includegraphics[width=1.0\linewidth]{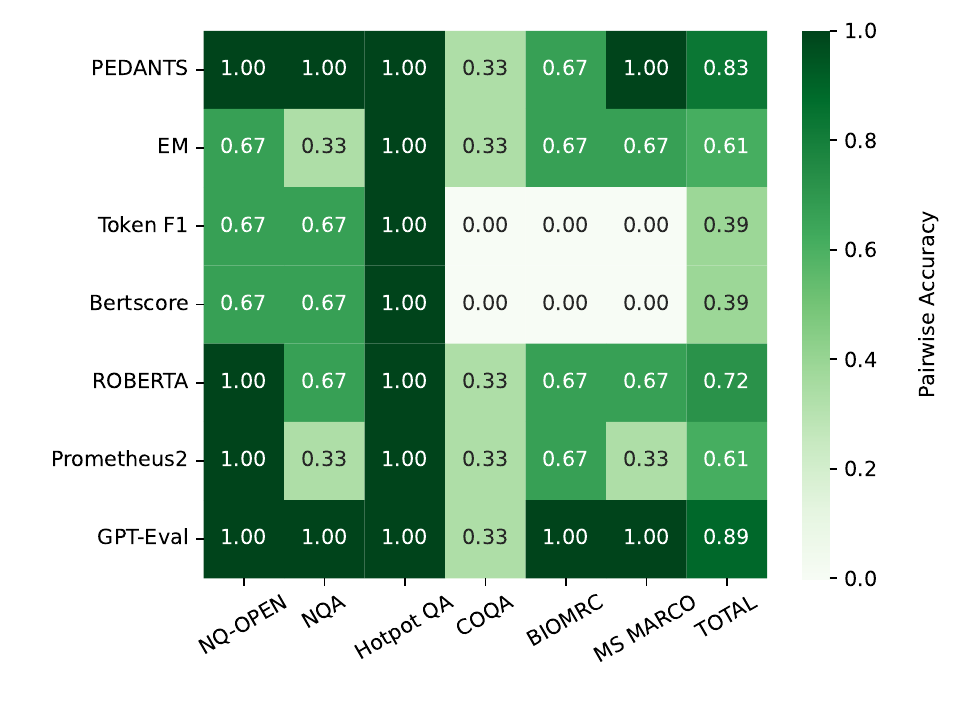}
    \caption{The pairwise ranking accuracy for dataset that have multiple model responses. The \textit{TOTAL} is the pairwise ranking accuracy across all six datasets. \cfm{} and GPT-Eval rank more models correctly than other methods on most datasets.}
    \label{fig:pairwise_acc}
\end{figure}

% And even the latency of local models matters
% if \ac{} must examine dozens of outputs from a process like beam search.
% %
% No evaluation metric is perfect in all aspects. 

% We use the trained $F(R)$ and $F(T)$ classifier to make rule and question type predictions to the 36,884 annotated examples and show the aggregated rule/question type distributions and \cfm{}'s human agreements in Figures~\ref{fig:aggregate_rule_distribution}, ~\ref{fig:aggregate_rule_accuracy}, ~\ref{fig:aggregate_type_distribution}, ~\ref{fig:aggregate_type_accuracy} (Appendix~\ref{sec:rule_human_agree_agrregate}). We do not see any examples of Rule 7 occurring in the test sets, suggesting that Rule 7 from our \ac{} rubric is not useful for our selected datasets. Additionally, Figure~\ref{fig:aggregate_rule_accuracy} shows that \cfm{} does the worst on Rule 2 (numerical information), and the best on Rule 1 (entity aliases). We analyzed 30 examples where \cfm{} gets wrong on Rule 2 examples, and we conclude that over 80\% of the examples are because \cfm{} have troubles understanding the conversion of data formats such as when reference is \textbf{\uline{Feb, 2018}} but the model generates \textit{\uline{$02/2018$}}, suggesting future improvement of \cfm{{} on detecting numerical and date information's correctness. Figure~\ref{fig:aggregate_type_accuracy} shows that \cfm{} is quite stable on various question types (no large variances among different types of questions), suggesting that the \textit{type} feature is less useful to align \cfm{} with human judgments than the \textit{rule} feature. .

\subsection{Error Analysis} 
We analyze failures of token, neural, prompt-based, and \cfm{} evaluations by manually analyzing 441 randomly sampled examples where they disagree with human judgments.
We categorize errors where these methods are likely to fail.
Then we examine what features contribute \cfm{}' most correctness judgments.

\subsection{Failures of Token and Neural Evaluations}
\label{subsection:human_eval}
We randomly select 63 examples from our annotated dataset where the methods---\abr{em}, token $F_1$, \bertscore{}, and \roberta{}, Prometheus 2, GPT-Eval, and \cfm{}—--show disagreement with human judgments. 
Two annotators manually analyze the 441 examples.
We discuss the failure cases separately within the contexts of token-level, neural, prompt-based, and \cfm{} evaluations.

\paragraph{Token evaluations are inflexible} 
Token evaluations makes more sense for \emph{extractive} \abr{qa}, where the goal is to find answers embedded in a fixed passage and compare token-level similarities of two strings.
However, generative models are capable of generating many valid and correct answers, making token evaluations less effective.
About 90\% (56) of the examples considered correct under human judgment are incorrect using \abr{em}. 
\abr{em} and token $F_1$ evaluations make the most mistakes on the following categories:
\begin{enumerate}
    \item \textit{Semantic similarity}: token evaluations struggle with catching semantic equivalence between two strings. For example, \textit{\refer{}: \uline{By labeling it satire} [SEP] \can{}: \uline{as being satirical}.} both convey the same meaning but \abr{em} and token $F_1$ will never judge them equivalent.
    \item \textit{Specificity}: this falls under \abr{ac} Rule 4, where \mm{}s often generate more detailed answers than gold answers. A representative example where \abr{em} and token $F_1$ both fail is \textit{question: The Aviation Hall of Fame and Museum of New Jersey is located at the airport in which New Jersey county? [SEP] \refer{}: \uline{Bergen} [SEP] \can{}: \uline{Bergen county, New Jersey}}.
\end{enumerate}

\paragraph{Neural evaluations cannot make sharp distinctions}
Unlike token-based methods, \bertscore{} and \roberta{} matching leverage pretrained or finetuned \abr{ac} knowledge and neural model representations, using \bert{}~\cite{devlin2018bert} to compute cosine similarity between the embeddings of the reference and candidate answers for evaluating correctness.
Because \roberta{} matching is finetuned specifically for \abr{ac} task, it gives  a more fine-grained evaluation than \bertscore{}.
However, one weakness of neural evaluation, even with finetuning, is its difficulty in distinguishing terms that are similar in the embedding space but have different meanings.
This weakness is especially obvious when the reference and candidate answers are short, where pretrained biases skew judgments.
Many of the incorrect answers that \bertscore{} and \roberta{} give high similarity scores and judge as equivalent are terms such as \textit{\uline{apple} and \uline{pear}; \uline{king} and \uline{queen}; \uline{left} and \uline{right}; \uline{item A} and \uline{item B}; etc}). 

\paragraph{Prompt-based evaluation decisions are unstable} Black box models and open-source \mm{}s have very different weaknesses.
Although black box models such as GPT-4 are strong evaluators, it is not sensitive to \textit{Specificity} due to word tokenizations during pretraining stage.
GPT-4 occasionally misjudges regarding the specificity of dates or locations. 
For example, \textit{\uline{Central Germany} and \uline{Western Europe}} are judged equivalent by GPT-4, where Western Europe can refer to many countries but Central Germany is only one country. 
\textit{Reference: \uline{March 25, 2018} and candidate: \uline{2018}} are also considered equivalent by GPT-4 when the question is asking for \textit{when} and the expected answer is expecting a specific date.

On the other hand, Prometheus is finetuned with mass synthetic evaluation data to evaluate the quality of long generations instead of just \ac{}, but it often misjudges simple short answers that are obviously wrong compared with the reference answers.
Prometheus often assigns a low rating when differences between the reference and candidate answers involve missing symbols, punctuation, or unnecessary omissions, such as \textit{reference: \uline{four albums}} and \textit{candidate: \uline{four}}. 
In this case, the question is \textit{How many albums had been recorded by Talking Heads by November, 1980?}, but Prometheus fails to consider the context of the question in its evaluation.
Prometheus also struggles at simple cases like whether \textit{\uline{Yes}} and \textit{\uline{No}} as equivalent, indicating that open-source \mm{}s may outperform standard evaluations like \abr{em} or \bertscore{} in evaluating long answers but are less effective at evaluating short-form factoid answers.

\paragraph{\cfm{} lacks world knowledge and commonsense reasoning} 
Although \cfm{} uses learned question types and \ac{} rules to judge \ac{} that goes beyond simple string matching, it is still a fairly cheap classifier with limited vocabularies.
From the 63 examples we analyzed, \cfm{} cannot handle complet word relationships but can be improved with future work. 
\textit{Synonyms that are not in the training vocabulary}: Unlike pretrained transformer models, \cfm{} lacks the general world knowledge to understand relationships between similar words. 
If the reference or candidate answer contains vocabularies that \cfm{} does not recognize such as \textit{\uline{cold} and \uline{chilly}}, \cfm{} becomes similar to token-based evaluation, which is a trade off for smaller model sizes, faster runtime, and less pretraining biases.
% \begin{enumerate}
    % \item \textit{Synonyms that are not in the training vocabulary}: Unlike pretrained transformer models, \cfm{} lacks the general world knowledge to understand relationships between similar words. If the reference or candidate answer contains vocabularies that \cfm{} does not recognize such as \textit{\uline{cold} and \uline{chilly}}, \cfm{} becomes a token-based evaluation, which is a trade off for smaller model sizes, faster runtime, and less pretraining biases.
    % \item \textit{Singular and Plurals}: without world knowledge about relationship between words, \cfm{} also struggles to judge singular and plural entities as equivalent such as \textit{\uline{apple} and \uline{apples}}, where \cfm{} treats them as two different entities.
% \end{enumerate}

\paragraph{Commonsense reasoning and fact-checking remain challenging tasks for all \abr{qa} evaluation} 
We analyze 45 examples from \jeopardy{} where all methods disagree with expert judgments. 
The 45 examples require more complex commonsense reasoning and fact-checking by human experts, which significantly challenge current \abr{qa} evaluation methods.
An interesting and challenging example is \textit{Question: Noted anarchist Prince Peter Alexeivitch Kropotkin wrote a 19th century entry on this capital? Reference: \uline{Moscow} Candidate: \uline{Saint Petersburg}}. 
The answer was also overruled to be correct by the panel with the fact that Kropotkin lived from 1842 to 1921; during his lifetime, the capital of \textit{\uline{Russia}} was changed from \textit{\uline{Saint Pertersburg}} (1712-1918) to \textit{\uline{Moscow}} (1918-). 
Thus, the capital Kropotkin referred to could be either of the answer.
%z
Experts assess correctness based on social norms, reasoning, and word choice, a process far more complex than mere string comparison.\footnote{See examples in Appendix~\ref{sec:pedanweaknessses}}
Evaluating a hard answer requires validating a fact and reasoning over a fact, if we want to adopt data from the expert community for evaluation, training, or analysis, we need to respect the community and also adopt their rules to improve \abr{qa} evaluations. 

\paragraph{\cfm{} Feature Importance} Understanding which features enable \cfm{} to make accurate judgment decisions can help us interpret its evaluation process and improve model transparency.
We use coefficient interpretation~\cite{hosmer2000applied} to identify the top five features that significantly influence \cfm{}'s decision-making: $R_1$, $F_1$ score, $T_{what}$, $R_3$, and tf-idf. 
These features show the critical role of \ac{} rules, question types, and $F_1$ scores as features in enhancing \cfm{}'s judgment accuracy, showing that a well-defined rules framework is important for \cfm{} to outperform traditional metrics like \abr{em}, $F_1$, and neural evaluation methods on our tests.

\label{error-analysis}

\label{analysis}

\section{Related Work}  

% \jbgcomment{This can be shortened considerably; just hit the high points and put the rest in the appendix}

\paragraph{Drawing on community expertise}

% \jbgcomment{I reworked this paragraph, would be good to find a better
%   citation for the ethical data use.  Anna Roger's ``dave'' paper is
%   the first one that comes to mind, but there might be better.}

There is a recent trend of using data from well-defined communities ethically~\cite{shilton2016emerging, rogers-etal-2021-just-think}, and the trivia community is no different.  While their data have been used for \abr{nlp} tasks---from
Watson's \textit{tour-de-force} on \textit{Jeopardy!} to
TriviaQA---the data have been used without attending to the norms of
how the trivia community uses it. We should listen to their evaluations if we adopt their data.

% As \citet{boyd-graber-borschinger-2020-question} argues, the goal of
% these competitions is to determine who is better at
% answering questions, and this is also the goal of modern leaderboards. adopting norms from the Trivia community can help judge \ac{}, which is a fiendishly hard problem for generative
% \abr{ai}.

% \jbgcomment{I think we can frame the ``listen to the community'' aspect of this a little more concretely: there are FATE publications about how you shouldn't just harvest data from a community without understanding and recognizing their values.  But this is exactly what TriviaQA, etc. have done.  Ben and I have a paper about how the norms of the trivia community should be adopted to improve QA, but this hasn't been put into practice.}

% \jbgcomment{I think we can talk about how QA evaluation sits on a continuum.  While many people recognize long-form eval is very difficult, some short-from QA is also hard.  It would be great to find examples that require translation, coreference, entailment, identifying metonomy, logical reasoning, etc.  This would make it clear that \equivalence{} is (or can be) a very hard problem that connects to many areas of \abr{nlp}.  Thus, an organization of this section could be:

%   1. many people recognize that \equivalence{} is hard in some cases (especially long-form qa)

%   2. but it can also be hard for short form qa, as recognized by the human qa community.  we should listen to them if we use their data

%   3. this is because even short-form qa eval can be nlp-complete: it contains ...

% }

\paragraph{Machine \abr{qa} evaluations}
\abr{qa} evaluations span a continuum—easy to assess using \abr{em} for short answers, yet challenging for long-form QA, which have many valid answers~\cite{Xu2023ACE}. However, even short-form QA can be challenging when it requires contextual understanding or have huge combination of valid output spaces\cite{bulian2022tomayto}. Such tasks can reach NLP-completeness, intersecting diverse NLP domains like coreference, translation, and entailment. 

% More details are available in Appendix~\ref{sec:answer_equivalence_in_other_fields}.
% \jbgcomment{ideally, these would be QA examples (hopefully mined from our experiments)}

% adopted rules from the Trivia community and an improved and more clear evaluation method are becoming more and more important than simple \abr{em} with continual improvement on transformer-base \abr{qa} models~\cite{qa_survey}. 
%
% Various benchmarks are proposed to evaluate the ability of \abr{qa} models to answer questions~\cite{benchmarks}. The benchmarks lead to improvement of \abr{qa} models, and improved \abr{qa} models can now answer questions much better by providing free-form answers or explanations that are more closely aligned with human preferences. 
%
With the continual development of machine \abr{qa} models' ability to answer questions, the area of \abr{qa} evaluation still falls behind, where most research papers still use standard evaluation metrics--\abr{em} or $F_1$ (Table~\ref{tab:qa_papers}). \citet{sameer-evaluating, kamalloo-etal-2023-evaluating} both point out pitfalls of current standard evaluation metrics, where \abr{qa} is not merely extracting exact answers from source texts in the era of large generative models. 
\citet{bulian2022tomayto, kamalloo-etal-2023-evaluating} validate fine-tuned \abr{bert} on human annotated \ac{} datasets is better than standard metrics and closer to human judgments. 
\citet{wang2023evaluating} demonstrates that \abr{bert}-based evaluation methods like \bertscore{} and \abr{blurt} \cite{sellam2020bleurt} are no better than standard metrics, with dataset sensitivity to \bertscore{} threshold settings and unclear boundary definitions across datasets, while generative \mm{} evaluators like \abr{gpt-3.5} tend to make \emph{over-generalization} errors that \abr{bert}-based methods and humans typically avoid.
Yet the ability of generative \mm{} evaluators is
not evaluated with clear definitions of \ac{} rigorously.

\label{related}

\section{Conclusion}

% \jbgcomment{I think we can/should also say that this could also be good for improving the models during training}
% \zongxiacomment{Feedback: people might argue that data-driven metric might not be interpretable than rule-based metrics. However, our construction of data is rule-based and the method composes a simple and interpretable classifier with additional features coming from current standard evaluation metrics, so it is more interpretable than pure data driven methods. Our LR-based method is more stable and has less variance on evaluating different models, which might have diverse styles of outputs. The method provides a more fair comparison between models since a change of response style does not affect the evaluation metric alignment with human judgment by a large margin.}

Automated \abr{qa} evaluation is an important pipeline for developing better models.
%
% A strict evaluation method such as \abr{em} provides too much negative feedback to models during training time, which results in guiding models to provide exact extracted answers. 
However, evaluation metrics like \abr{em} are popular for their efficiency, but the lack of \ac{} rubrics and diverse evaluation datasets still hold us from exploring and building more robust and stable \abr{qa} metrics competitive with \mm{} evaluations, which is not always available.
A basic beginning framework derived from
\abr{qa} human experts can enhance and generalize automated evaluation methods' agreement more with human judgments.
Though not perfect, \cfm{} requires few computational resources and time but more robust across benchmark datasets. 
%
% A better evaluation method during training time can improve models' downstream task accuracy ~\cite{answerExpansion}. 
%
Future work can integrate and refine these rubrics into long-form \abr{qa} and novel \abr{qa}, where we can collect better evaluation data and improve fact-checking and commonsense reasoning of \ac{}; future work can also combine efficient metric with \mm{} to improve runtime evaluation efficiency and reduce cost.

\label{conclusion}

\section*{Limitations}
Since we are the first to adopt and revise answer correctness rules from the Trivia community, they are still not the perfect within the machine evaluation paradigm.
We advocate stronger connections between the computer linguistic community and the professional Trivia \abr{qa} community to improve the interpretability, efficiency, and effectiveness of rule-based \abr{qa} evaluation.
Furthermore, although \cfm{} has shown its efficiency and effectiveness on standard \abr{qa} benchmark datasets than token evaluation and neural evaluations, it still has drawbacks that need to be address.
The effectiveness of \cfm{} is constrained by its limited vocabulary. 
When encountering words outside its training corpus, \cfm{} becomes to the less-effective token-level evaluation. 
This vocabulary limitation also impedes \cfm{}'s capacity to capture complex word associations, thereby weakening its commonsense reasoning abilities.
Future research is needed for expanding \cfm{}'s vocabulary range and improve its ability to learn complex relationships between words. 
Such improvements can bridge the gap between efficient \cfm{} evaluation and expert human evaluations.

% \cfm{} trained with \ac{} data containing our framework examples can better justify the correctness of candidate answers given a reference answer than current popular \abr{em}, token $F_1$ and neural evaluation methods. However, our new method still possesses several challenges and limitations. Our models are not trained to take the contexts of the question into account, whereas if a question has a context paragraph, our model can only use the question, reference answer, and candidate answer to evaluate correctness, which the effectiveness of our approach is still unexplored for \abr{qa} with contexts. Another limitation is the lack of validation of fact-checking and commonsense-reasoning of \abr{qa} evaluations, which is also not validated against \abr{gpt-4}. In addition, our definition of \ac{} has not taken subjectivity, demographic diversity, bias, and cultural diversity into account, with cases where the judgments of correctness can be different for people with different backgrounds, where we consider such subjective questions as maximal difficult and challenge examples for \ac{} tasks. 
\label{limitation}

\section*{Ethics}
% Our annotations do not potent any risks to participants. We do not collect any personal information from participants. Participants are allowed to participate or exit the study reading our task instructions within ten minutes after they started the study by any means. All of our annotations are supervised by the Institutional Review Board, which judged the annotation to be exempt. Each annotator is compensated with \$7.5 dollars half an hour rate to complete 300 examples. Annotators will have a fixed rate if they completed 300 examples or a bonus with 2.5 cents with each additional example for each additional examples above 300 examples. Our python package is open-source and protected under the MIT license. We welcome public download and usage of our \abr{qa} evaluation python package for any purposes. 

Our annotation process prioritizes participants' privacy. 
We do not collect any personal information from participants, and they are free to exit the study within ten minutes of starting. 
The Institutional Review Board (\abr{irb}) has reviewed and exempted our annotation protocol.
Annotators are compensated at a rate of \$15 per hour, with an expected completion of 300 examples in this time frame. 
For productivity beyond 300 examples, we offer a bonus of 2.5 cents per additional example.
Our open-source QA evaluation Python package is released under the MIT license, encouraging public use and contribution. 
We welcome its application for any purpose, supporting transparency and advancement in the field.

\label{ethics}

\section*{Acknowledgement}
We thank the invaluable contributions from University of Maryland CLIP members who initiated this project through their insightful brainstorming. 
We extend our gratitude to the annotators for their efforts in annotating the data and for Jonah Greenthal for his discussions of answer equivalence.
Additionally, we appreciate the anonymous reviewers for their constructive feedback, which has significantly enhanced the robustness and rigor of our \abr{qa} evaluation and analysis. 
The improvements in \cfm{} owe much to their valuable comments and suggestions.
This material is based upon work supported by the National Science
Foundation under Grant No. \abr{iis}-2403436 (Boyd-Graber).
Any opinions, findings, and conclusions or recommendations expressed
in this material are those of the author(s) and do not necessarily
reflect the views of the National Science Foundation.
\label{acknowledgement}

\bibliography{bib/journal-full,bib/custom,bib/jbg}
\bibliographystyle{style/acl_natbib}

\appendix
\begin{table*}[ht]
\centering
\small
\begin{tabular}{m{0.139\textwidth} m{0.139\textwidth} m{0.139\textwidth} m{0.139\textwidth} m{0.139\textwidth}}
\hline
\textbf{Source} & \textbf{Year} & \textbf{Model} & \textbf{Training Metric} & \textbf{Datasets} \\ \hline
\citet{ainslie2023gqa} & 2023 & MHA-Large, MHA-XXL, MQA-XXL-GQA-8-XXL & \abr{bleu}, $F_1$ & CNN/Daily Mail, arXiv and PubMed, MediaSum, Multi-News, TriviaQA \\ \hline 
\citet{jiang2023active} & 2023 & GPT-3.5, text-davinci-003 & \abr{em}, $F_1$ & 2WikiMultihopQA, StrategyQA, ASQA, ASQA-hint, WikiAsp \\ \hline
\citet{liu2023lost} & 2023 & LongChat-13B, MPT-30B-Instruct, GPT-4.5-Turbo, Claude-1.3, Flan-T5 xxl\dots & \abr{em} & \abr{nq-open} \\ \hline
\citet{bevilacqua2022autoregressive} & 2022 & DPR, BM25, GAR, DSI-BART, SEAL (BART-Large) & \abr{em} & \abr{nq-open} 
\\ \hline
\citet{zhang2023merging} & 2023 & DPR (FiD and RoBERTa-large), GPT-3.5-turbo & \abr{em} & \abr{nq-open}, TriviaQA, WebQuestion, HotpotQA 
\\ \hline
\citet{jeong2024adaptiverag} & 2024 & Flan-T5 xl (3B), Flan-T5 xxl (11B), GPT-3.5-Turbo & \abr{em}, $F_1$ & \abr{nq-open}, TriviaQA, SQuAD
\\ \hline
\citet{chen2024improving} & 2024 &  Electra Small Discriminator & \abr{em}, $F_1$ & Adversarial SQuAD, TriviaQA, SQuAD
\\ \hline
\citet{zhu2023chainofquestions} & 2024 &  BERT, GPT-3.5, T5-B, T5-L, TB-T5-L &  $F_1$ & HotpotQA, DROP~\cite{dua-etal-2019-drop}
\\ \hline
\citet{zhu2023chainofquestions} & 2022 &  RoBERTA, BART, ELECTRA~\cite{clark2020electra}, T5 & \abr{em}, $F_1$ & SQuAD, NewsQA, TriviaQA, SearchQA, HotpotQA, \abr{nq-open}, DROP, RACE\dots
\\ \hline
\citet{luo2023sail} & 2023 &  LLaMA (7B), Vicuna (13B), ChatGPT & GPT-4-Eval & CommonsenseQA~\cite{talmor-etal-2019-commonsenseqa}, OpenbookQA~\cite{mihaylov-etal-2018-suit}, ARC-Challenge~\cite{clark2018think}
\\ \hline
\citet{liu2024chatqa} & 2024 &  GPT-3.5, GPT-4, Llama3-ChatQA & \abr{em}, $F_1$ & COQA~\cite{reddy2019coqa}, ChatRAG BENCH~\cite{liu2024chatqa}
\\ \hline
\citet{xu2024debateqaevaluatingquestionanswering} & 2024 &  Gemma, LLama3 GPT-4, Phi-3 & \abr{em}, $F_1$ & \cite{xu2024debateqaevaluatingquestionanswering}
\\ \hline
\citet{yona-etal-2024-narrowing} & 2024 &  PaLM 2 & BLEURT & GRANOLA \cite{yona-etal-2024-narrowing}
\\ \hline
\citet{meta_llama_2024} & 2024 & LLaMA-3.1-8B & \abr{em}, $F_1$, accuracy, pass@1 (coding) & 
Llama-3.1-8B-evals
\\ \hline
\end{tabular} 
\caption{Question answering model training papers and the evaluation metrics used during training. In the year of 2024, most of the \abr{qa} training papers still use \abr{em} and $F_1$ as the evaluation metrics.
% The \jeopardy{} dataset also includes the expert human responses and judgments of correctness.
}
\label{tab:qa_papers}
\end{table*}

\section{\ac{} and Correctness Guideline Examples}

\begin{figure*}[!t]
    \makebox[\linewidth][c]{%
        % \hspace{-0.5cm}  % Adjust X to the desired value
        \includegraphics[scale=1.00]{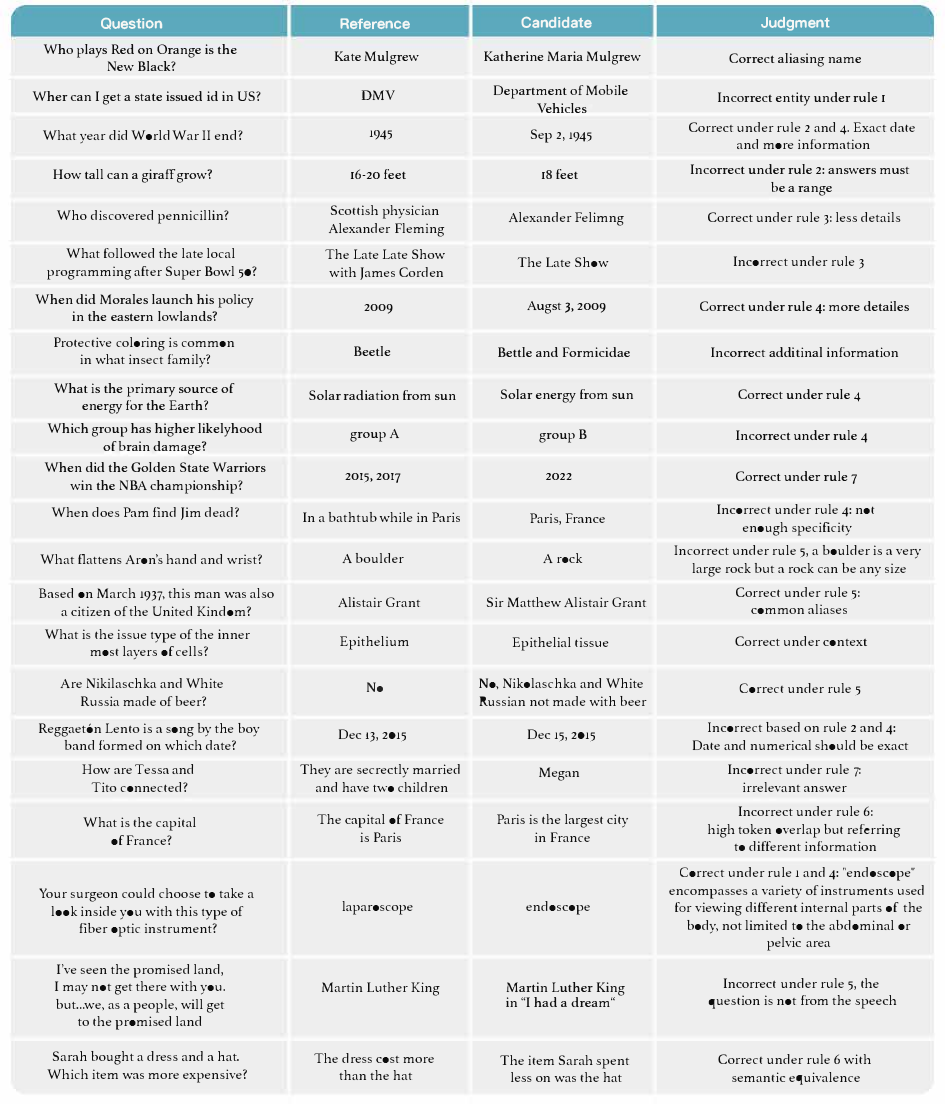}
    }
    \caption{We list out more relevant \ac{} pairs with judgments under our revised rules. Some of the \can{} responses are manually written, some of them are from \jeopardy{}, and some of them are generated by various \abr{qa} models such as Flan-t5.}
    \label{tab:more_ae_examples}
\end{figure*}

% \begin{figure*}[!t]
%     \makebox[\linewidth][c]{%
%         \hspace{-0.5cm}  % Adjust X to the desired value
%         \includegraphics[scale=0.45]{figures/gpt_comparison3.pdf}
%     }
%     \caption{Classifier-based evaluation methods with augmented \ac{} training set are more stable than standard evaluation methods on different models. In addition, \abr{lr}-based methods trained on our augmented dataset is close to or even better than \abr{bert}-based methods' judgment alignment with our fine-tuned \ac{} \abr{gpt-3.5-turbo} with constantly at least $80\%$ of alignment. On the other hand, standard evaluation methods are unstable with the change of \abr{qa} models or datasets. The error bars are the normalized variances of alignment percentage for the three \abr{qa} models under the same dataset.}
%     \label{fig:gpt_alignment}
% \end{figure*}

\label{section:more_ae_examples}
Table~\ref{tab:more_ae_examples} presents additional typical \abr{qa} example pairs, along with corresponding judgments and explanations. 
These examples provide clarity on how rules in Table~\ref{tab:guidelines} are applied.

\section{Training Data Augmentation}
\label{sec:data_augmentation}
Since no existing \ac{} datasets are built according to our rules, we will generate additional synthetic data and augment the existing \ac{} dataset from \citet{bulian2022tomayto}. 
By combining this synthetic data with real human-annotated examples, we aim to enhance \cfm{}' performance by distilling knowledge from GPT-4~\cite{zhou2024teachingassistantintheloopimprovingknowledgedistillation}.

We first build representative \abr{qa} example pairs that align with the revised \ac{} and acceptability rules outlined in Table~\ref{tab:guidelines}, and train a classifier to learn the optimal threshold based on these patterns. 
For each \ac{} rule, we gather a small set of \abr{qa} pairs from the \abr{naqt} correctness rubrics and error examples from the \ac{} test set and the \abr{ropes} \textit{dev} set. We then manually revise \abr{qa} pairs that violate our \ac{} rubrics to ensure both quality and diversity. For each rule, we create five to ten examples, balancing the number of correct and incorrect pairs, except for rules that are strictly incorrect.

\paragraph{Seed Examples} Each rule is composed of 5 to 10 representative seed examples,where each example is strictly judged based with our rules. However, we do not guarantee that each generated example only corresponds to one rule (may correspond to multiple rules). An example seed example for rule 1 is \textit{question: In 2011, what airport did the most international travelers in North America visit? [SEP] reference: John F. Kennedy International Airport [SEP] candidate: JFK Airport}.

\paragraph{Training data generation}
We use the manually revised \abr{qa} pairs as seed examples to prompt
\abr{gpt-4} with an example and the specific \ac{} rule to generate
more similar \abr{qa} pairs with its judgment of equivalency and
correctness.\footnote{We set the temperature of \abr{gpt-4} to 0.9 and
frequency penalty to 1.9 to ensure the diversity of generated \abr{qa}
pairs.} At the end of each generation iteration, we append the
generated examples to the seed examples, which we generate a total of
4,234 synthetic \ac{} examples.

\paragraph{Quality validation}
To verify the automatically generated examples, we manually select fifty generated \ac{} examples and check the judgments.
\abr{gpt-4}'s disagreements with humans is most frequent with Rule~2
from Table~\ref{tab:guidelines}, e.g., \abr{gpt-4} often considers
numerical answers like \textit{\uline{2008-2010} and \uline{2010}} to be equivalent.
For each generated example, we run \abr{gpt-4} self-evaluation by prompting it to verify the correctness of its generated judgments with the rules, questions, \refer{} answers, and \can{} answer, and judgment to improved the quality and accuracy of generations~\cite{ren2023selfevaluation, zhou2024teachingassistantintheloopimprovingknowledgedistillation}.

% To improve the quality of the generated judgements, we run \abr{gpt-4}
% again on the generated examples by prompting it with the rules,
% questions, \refer{} answers, and \can{} answers, which can
% improved the quality and accuracy of generated
% contents~\cite{ren2023selfevaluation}.
% \jbgcomment{Previous sentence is unclear, perhaps give an example}
%
We re-check the 50 generated examples that have inconsistent generated
and self-evaluated judgments, and see an improvement of the
judgments---39 out of 50 example are consistent with human judgments
after self-evaluation. Moreover, we select 300 additional generated
examples (mainly examples on Rule 2) and manually check and revise the
self-corrected judgments to accommodate our \ac{} rubrics. 180
self-verified judgments are revised and 65 invalid examples (false
\refer{} answers, invalid questions) are removed with 3,989 examples
remaining in our training material. After two steps of validation, we
are confident that the synthetic data mostly align well with our \ac{} rules.

\paragraph{Extra Human Annotated Data Augmentation}
The first step of data augmentation is to collect representative
materials to fit the classifier to follow \ac{} norms.
We also
want to generalize and teach the classifier with more real human
annotated judgments. Luckily, we already an existing annotated
\ac{} dataset available. We expand the \ac{} learning material
by combining our 3,989 synthetic examples with 9,090 examples from the
\ac{} training set. 

\section{Model Training Details}
\label{sec:traing_hyperparameters}
We provide specific training details for logistic classifier and fine-tuning details for \roberta{} for ease of reproduction. 

%
% \paragraph{\abr{bert}-Tiny training} We train \abr{bert}-Tiny with objective function~\ref{eq:unified_contrastive_loss}, learning rate = \textit{1e-5}, weight decay=\textit{0.01}, batch size= \textit{16} for 20 epochs and save the model with the highest accuracy.

%
\paragraph{Logistic classifier} We use \texttt{random\_state}=\textit{668}, \texttt{loss}=\textit{log\_loss}, \texttt{penalty}=\textit{l2}, \texttt{tol}=\textit{1e-3}.

\paragraph{\roberta{} fine-tuning} We trained \roberta{}-base with learning rate = \textit{1e-5}, weight decay=\textit{0.01}, batch size=\textit{32} for 2 epochs and save the state with highest accuracy on the \ac{} test dataset following \citet{bulian2022tomayto}'s input training format.

\section{Evaluation Dataset Details}
\label{sec:eval_data_details}
Table~\ref{tab:dataset} shows examples and number of \abr{qa} pairs for each of our selected benchmark dataset. \abr{nq-open}, Hotpot\abr{qa}, and Narrative-\abr{qa} are short-form \abr{qa}, where the reference answers are usually very short, and there are multiple correct reference answers. \abr{coqa}~\cite{reddy2019coqa} is a large-scale dataset designed to assess machine comprehension of text through a conversational question and answer format, featuring over 127,000 questions derived from 8,000+ dialogues across seven diverse domains. Each question in CoQA is part of a conversation about a passage, with answers provided in free-form text alongside evidence from the text, challenging systems with tasks like coreference and pragmatic reasoning. \abr{biomrc}~\cite{stavropoulos2020biomrc} is a biomedical machine reading comprehension task dataset that represent real-world machine biomedical reading comprehension tasks. \abr{ms marco} is a question answering dataset of 100,000 real Bing questions, each paired with a human-generated answer. For each of these datasets, we choose a subset by random sampling for evaluation except for \abr{coqa}, where each conversation contains 10 consecutive questions. We randomly sample 300 conversations from \abr{coqa}. 

\begin{table*}[ht]
\centering
\small
\begin{tabular}{m{0.13\textwidth} m{0.08\textwidth} m{0.30\textwidth} m{0.18\textwidth} m{0.16\textwidth}}
\hline
\textbf{Dataset} & \textbf{\# Pairs} & \textbf{Context} & \textbf{Question} & \textbf{Gold Answer} \\ \hline
% \abr{\abr{ropes}} & 1,688 & Michael Paul Welch (born August 25, 1972) is a former right-handed pitcher ...  was named a 2014 "Best High School" by U.S. News and World Report. & Mike Welch attended high school that serves how many students ? & 2,200 students \\ \hline
\abr{nq-open} & 3,610 & \multicolumn{1}{c}{-} & who won the American league east in 2017 & The Yankees, Houston Astros \\ \hline 
Hotpot\abr{qa} & 3,300 & Asmara International Airport (IATA: ASM) (ICAO: HHAS). Asmara currently hosts the country's only operating international airport... & Asmara international airport is in which country? & Eritrea \\ \hline
Narrative-\abr{qa} & 3,300 & The narrator, a Bostonian, returns after a brief visit a few summers prior, to the small coastal town ... cMaine coastal town increases each day. & What was Littlepage's job? & sailor, retired sailor \\ \hline
\abr{coqa} & 5,500 & CHAPTER XXX. THE MEETING FOR RAIN. Meanwhile the Auld Lichts were in church, waiting for their minister... & Who gaped at Hendry? & Peter was so taken aback that he merely gaped at Hendry \\ \hline
\abr{biomrc} & 1,373 & Why is a second dose of MMR necessary? About $2\%-5\%$ of persons do not develop measles immunity after the first dose of vaccine. This occurs for a variety of reasons. The second dose is to provide... & Why is a second dose of mmr necessary & To provide another chance to develop measles immunity for persons who did not respond to the first dose. \\ \hline
\abr{ms marco} & 3,000 & It is located on a 100-acre (0.40 km 2) site on the northern quay of the Royal Victoria Dock in London Docklands, between Canary Wharf and London City Airport. Phase II was... & where is excel arena london & It is located on a 100-acre (0.40 km 2) site on the northern quay of the Royal Victoria Dock in London Docklands, between Canary Wharf and London City Airport. \\ \hline
\abr{MOCHA} & 5,033 & Somewhere in me I knew it all along, there are all those moments when he stares into my eyes and his start to sparkle while this gorgeous grin spreads across his face... & What's a possible reason the guy stares into the writer's eyes ? & Because he likes her a lot/He's a child and it's a very rare thing. \\ 
% \hline
% \jeopardy{} & 371 & \multicolumn{1}{c}{-} & During the war, this Ozark state had 2 competing governments? & Arkansas \\ 
% \abr{DEBATEQA} & 5,033 & Somewhere in me I knew it all along , there are all those moments when he stares into my eyes and his start to sparkle while this gorgeous grin spreads across his face... & What's a possible reason the guy stares into the writer's eyes ? & Because he likes her a lot/He's a child and it's a very rare thing. \\ 
\hline
\end{tabular}
\caption{The seven datasets we use for our generalization testing. The later three datasets generally have much longer gold answers than the previous three, which are considered more challenging. We do not include contexts for \abr{nq-open} for more variability of generated answers to challenge the evaluation methods. MOCHA already includes human annotated labels.
% The \jeopardy{} dataset also includes the expert human responses and judgments of correctness.
}

% \caption{Number of \abr{qa} pair examples and datasets we use in our study to evaluate automated \abr{qa} evaluation metrics. The cell with `-' indicates the \abr{qa} pairs do not have related contexts.}

\label{tab:dataset}
\end{table*}

\section{$F_1$ Score Details}
\label{appendix:F1_details}
We provide additional formulas as details to calculate the precision, recall, and $F_1$ score between a reference and a candidate answer. 
In the context of evaluating a candidate answer against a reference answer, precision, recall, and $F_1$ score determine the extent of two strings' similarity by tokenizing both answers into individual words. 
We define tokens as words separated by any white spaces or tabs, where a string \textit{s} can be split to a list of tokens \textit{Tok(s)}. 
Precision \textit{P} is the ratio of the number of tokens that are present in both the candidate and the reference answer to the total number of tokens in the candidate answer. 
\textit{P} evaluates the accuracy of the candidate answer by indicating the proportion of its tokens that are relevant to the gold answer:
\begin{equation}
\textit{P} = \frac{|\textit{Tok}(\textit{candidate}) \cap \textit{Tok}(\textit{reference})|}{|\textit{Tok}(\textit{candidate})|}
\label{eq:precision}
\end{equation}

Recall \textit{R} measures the proportion of tokens from the reference answer that are captured by the candidate answer, providing insight into the completeness of the candidate's response:
\begin{equation}
\textit{R} = \frac{|\textit{Tok}(\textit{candidate}) \cap \textit{Tok}(\textit{reference})|}{|\textit{Tok}(\textit{reference})|}
\label{eq:recall}
\end{equation}
% \[
% R = \frac{| \text{{Tokens in both candidate and gold answer}} |}{| \text{{Tokens in gold answer}} |} \label{eq:recall}
% \]
$F_1$ score is the harmonic mean of precision and recall, which balances between the precision and recall metrics. It is particularly useful when the importance of false positives and false negatives are equally significant. The F1 score is computed as follows:
\begin{equation}
F_1 = 2 \times \frac{\textit{P} \times \textit{R}}{\textit{P} + \textit{R}} \label{eq:f1score}
\end{equation}
The three metrics provide a comprehensive token matching evaluation of the candidate answer's relevance and completeness concerning the gold answer, and are used in many \abr{qa} model training/evaluation practices mentioned in Table~\ref{tab:qa_papers}.

\section{\jeopardy{} Data Collection Pipeline} 
\label{appendix:jeopardy}
The
\textit{J!Archive} is a fans website to record every season of the
\jeopardy{} game with questions answers
presented.\footnote{\url{https://www.j-archive.com/suggestcorrection.php?clue_id=353154}}
The website also includes questions and answers that are answered by
\jeopardy{} players that are originally judged
correct/incorrect, but with later revised decisions, where those
overruled answers are carefully verified and discussed by audiences or
experts. We collected a total of 504 (half correct, half incorrect) such examples from J!Archive use
them as the challenge set.

\section{Comparing Expert \equivalence{} Judgements}
\label{sec:appendix_Jeopardy}
The \jeopardy{} dataset is a challenge test set that has a very different source of answers with expert judgments and use it to evaluate the effectiveness of various \abr{qa} evaluation methods on the long-long tail of \abr{qa}. We introduce this \abr{ood} challenge test set that has a very different source of answers with expert judgments and use it evaluate the effectiveness of various \abr{qa} evaluation methods.
\paragraph{\jeopardy{}} is a popular American television quiz game show. This show is famous for its distinctive formats of questions, where the questions covers diverse categories, and many of them require a knowledgeable human with years of expertise to be able to answer. An important note is that that there is a pipeline of people who have worked for \abr{naqt} have gone on to contribute to \jeopardy{} in one way or another, the most notable of which is Ken Jennings, who became the first response adjudicator for
\abr{naqt}~\cite{ken_jenning}. Thus the selected \jeopardy{}
challenge problems also conform to our \ac{} rules, where we
will remove examples with rules we did not adopt (pronunciation,
name orders, etc).~\footnote{\jeopardy{} calls questions
clues rather than questions, and the answers are responses. We
rephrase the clues to be questions and response to gold answers to
conform to machine \abr{qa} format.}

\paragraph{Expert dataset format} %
% \jbgcomment{I wonder if this should come later when we talk about Jeopardy in more detail, as is we're introducing Jeopardy! twice} 
Technically, the responses to \jeopardy{} clues are ``questions'' that provide the information sought by the ``answers'' asked by the host, a gimmick in response to the 
crucible of the mid-century game show crisis~\cite{stone1992prime}.  Although treating \uline{Who is Cleopatra?} as a full answer to a question would be a further challenge to \ac{} systems, we treat \textit{Jeopardy!} responses as everything after the verb; in this case, \uline{Cleopatra}.

\paragraph{Expert dataset details} We collect 504 examples from \textit{J!Archive}, a fans website to record every season of the \jeopardy{} game with questions and answers presented.\footnote{\url{https://www.j-archive.com/suggestcorrection.php?clue_id=353154}} Each example in the dataset has a question, a \refer{} answer, a \can{} response by different \jeopardy{} players, and expert judgment of the \can{} answer. The dataset includes both difficult correct answers and difficult incorrect answer. E.g., the difficult correct answers are the ones the expert host ruled incorrect but was overruled by a panel. For example, given a question \textit{I've seen the promised land, I may not get there with you, but...we, as a people, will get to the promised land}, the \refer{} answer is \uline{Martin Luther King}, but \uline{Martin Luther King in "I had a dream"} was given and ruled correct during the show that later reverted to be incorrect, verified by professionals in accordance with the \jeopardy{} answer acceptability rules,  as the speech in question does not originate from the "I Have a Dream" speech. 
The dataset includes 50\% correct examples and 50\% incorrect examples, which we use to test the long-tail automated \abr{qa} evaluation methods. 

% \subsection{Is \abr{qa} Evaluation Solved\dots?}
% \label{sec:Jeopardy}
% While our evaluation works \emph{now}, we recognize that it is not rich enough for all \ac{} tasks.
%

\subsection{Human Annotations}
\label{sec:human_annotation}
% Each metric evaluates outputs from six models for each dataset (except \abr{mocha}). We calculate
% the agreement accuracy and Macro $F_1$ with human judgments for each method
% (Table~\ref{tab:human alignment}) and analyze \cfm{} weaknesses in
% Section~\ref{analysis}.
% and we calculate the standard deviation of human agreement percentage across models for each metric. 
%
%
% We annotate random sample of 
% %
% 20,258 examples from \abr{nq-open}, \abr{nqa}, and Hotpot \abr{qa} and 14,000 examples on the three long answer datasets. 
% We hire annotators on Prolific whose first language is English, with at least
% a community college degree completed and above 99\% approval rate.
% %
% Annotators use our instructions of \ac{} (Table~\ref{tab:guidelines}) and we obtained 
% 36,884 annotations, where 6,626 examples have two annotations. 801 (12.1\%)
% annotations disagree (Krippendorff's $\alpha=0.75$), and authors break
% ties.\footnote{Exact details of the number of annotations for each dataset is
%   in Appendix~\ref{sec:annotation}}
We delve into details on human annotations in this section. We hired crowdsource annotators from prolific with a 99\% approval rate. Each annotator is presented with a set of rules from Table~\ref{tab:guidelines}. 
Each rule has an example to justify the judgment shown in Table~\ref{table:generation_prompt_template}. For the three datasets with short-form answers -- \abr{nq-open}, \abr{nqa}, \abr{hotpot-qa}, we select 17,602 examples \abr{em} judged as incorrect. Specifically, among the 17,272 examples, 6,626 of them are annotated by two different annotators. Among the 6,626 examples with double annotations,  801 (12.1\%)
annotations disagree (Krippendorff's $\alpha=0.75$), and we need manually go over those examples to select the better judgment. Thus, the three short-form \abr{qa} datasets give us 17,272+6,626 = 23,898 examples. 

In addition, we select 24,488 examples from the challenge datasets that do not have an exact match with longer reference answers and hired annotators. Thus, our total number of annotations is 23,898+24,488=48,386.

\begin{table*}[h!]
\centering
\small
\begin{tabular}
{>{\raggedright\arraybackslash}p{0.95\linewidth}}
\hline
\textbf{Prompt template for generating \equivalence{} examples} \\
\hline
You will be provided with a specific rule to assess the correctness of a candidate answer given a reference answer. Your assignment involves generating high quality diverse example questions different from the given example. Each example must include a question, a reference answer, 
a candidate answer, and a judgment on whether the candidate answer is correct based on the provided rule. 
It is important that your examples follow the structure of the given example, with an emphasis on ensuring 
that the reference and candidate answers are similar but not identical. \\ \\
\textbf{[Example]} \\ 
Rule: \textcolor{red}{Selected Rule}\\
\textcolor{red}{Your Seed Example}
\\ \\
\textbf{[Instruction]}\\
1. Try to include hard negative and hard positive examples in your responses.\\
2. Try to make your responses as diverse as possible.\\ 
3. Only generate json format responses.\\ \\
Rule: \textcolor{red}{Selected Rule}\\
\textbf{[Your Response]}\\
\hline
\end{tabular}
\caption{The prompt template for generating \equivalence{} examples. The red texts are your inputs.}
\label{table:generation_prompt_template}
\end{table*}

\begin{table*}[h!]
\centering
\small
\begin{tabular}{>{\raggedright\arraybackslash}p{0.95\linewidth}}
\hline
\textbf{Prompt template for self-verifying} \\
\hline
You are given a correctness rule, a question, a reference answer, and a candidate answer to the question. Your task is to determine the correctness of the candidate answer by outputting either \textbf{correct} or \textbf{incorrect}. \\ \\
Rule: \textcolor{red}{Corresponding Rule}\\ \\
\textbf{[Example]} \\
Question: \textcolor{red}{Example question}\\
Reference: \textcolor{red}{Example reference answer}\\
Candidate: \textcolor{red}{Example candidate answer} \\
Judgment: \textcolor{red}{Gold judgment}\\ \\
\textbf{[Input]}\\
Question: \textcolor{red}{Example question}\\
Reference: \textcolor{red}{Example reference answer}\\
Candidate: \textcolor{red}{Example candidate answer} \\
Judgment: 
\\
\hline
\end{tabular}
\caption{The prompt template for self-verifying the generated examples.}
\label{table:self_verification}
\end{table*}

\begin{table*}[h!]
\centering
\small
\begin{tabular}{>{\raggedright\arraybackslash}p{0.95\linewidth}}
\hline
\textbf{Prompt template for \abr{gpt-4}-Eval} \\
\hline
\textbf{[1]} You will be given a set of rules and representative examples for each rule to determine whether the \can{} is correct based on the question and the \refer{} answer.\\ 
\textbf{[2]} Then you will be given a question, a \refer{} answer, and a \can{} answer. You tasks is to determine whether the \can{} is correct based on the question, the \refer{} answer, and the rules and examples. Please only output the rule number and \textbf{"correct"} or \textbf{"incorrect"} as the output.\\ \\
\textbf{Rule 1}: Widely recognized aliases, pseudonyms, or alternative names that are commonly associated with the \refer{} answer entities are acceptable.\\
\textbf{Example 1}: Question: Who plays Red on Orange is the New Black? \refer{} Answer: Kate Mulgrew. \can{} Answer: Katherine Maria Mulgrew. Your Answer: Rule 1,correct.\\
\textbf{Rule 2}: Exact dates, years, numerical are required unless the question specifically asks for approximations. \\
\textbf{Example 2}: Question: How tall can a giraffe grow? \refer{}: 16-20 feet. \can{}: 18 feet. Your Answer: Rule2,incorrect.\\
\textbf{Rule 3}: The \can{} answer provides less details but should include the essential and correct information required by the question (specificity 1).\\
\textbf{Example 3}: Question: What followed the late local programming after Super Bowl 50? \refer{}: The Late Late Show with James Corden. \can{}: The Late Show. Your Answer: Rule 3,incorrect.\\
\textbf{Rule 4}: \can{} answer contains the \refer{} answer or an equivalent part. The additional information must be factually correct and does not contradict the question or \refer{} answer (Specificity 2).\\
\textbf{Example 4}: Question: When did Morales launch his policy in the eastern lowlands? \refer{}: 2009. \can{}: August 3, 2009. Your Answer: Rule4,correct.\\
\textbf{Rule 5}: A high degree of word overlap or similarity does not establish equivalence. The answer must be contextually and semantically accurate.\\
\textbf{Example 5}: Question: What is the primary source of energy for the Earth? \refer{}: Solar radiation from sun. \can{}: Solar energy from sun. Your Answer: Rule 5,incorrect.\\
\textbf{Rule 6}: The \can{} answer is irrelevant to question and the given answer or is an inaccurate description of the question should be incorrect.\\
\textbf{Example 6}: Question: Which event occurred first, the first Super Bowl or the formation of the Premier League? \refer{}: first Super Bowl. \can{}: formation of the Premier League. Your Answer: Rule 6,incorrect.\\
\textbf{Rule 7}: \can{} response is correct but not in the \refer{} list.\\
\textbf{Example 7}: Question: When did the Golden State Warriors win the NBA Championship? \refer{}: 2015, 2017. \can{}: 2016. Your Answer: Rule 7,correct.\\
Please output your answer below.\\
Question: \textcolor{red}{\textbf{[Input question]}}\\
\refer{}: \textcolor{red}{\textbf{[Input reference answer]}} \\
\can{}: \textcolor{red}{\textbf{[Input candidate answer]}}\\
Your Answer: \\
\hline
\end{tabular}
\caption{The prompt template for \abr{gpt-4}-Eval. We include representative examples for each rule to align \abr{gpt-4}'s judgment with the rules.}
\label{table:eval_prompt_template}
\end{table*}

\section{\ac{} in Other Fields}
\label{sec:answer_equivalence_in_other_fields}
\ac{} connects many areas of \abr{nlp}, where many of the following tasks can be considered a variation of answer correctness.

\paragraph{Coreference} refers to when two or more expressions in a text refer to the same entity, e.g., person~\cite{an2023nichellenancyinfluencedemographic}, place, thing~\cite{wu-etal-2020-corefqa}. Equivalence examples are \textit{\uline{Her boyfriend} and \uline{Kristy's boyfriend}} given question \textit{Who showed up to help Kristy escape when the house was collapsing?}. 
Since the given question already mentions the name Kristy, \textit{her} in the answer also refers to Kristy's.
\paragraph{Translation} involves converting text or speech from one language to another~\cite{bahdanau2016neural}. Generative models are also likely to provide answers not just in one language. \textit{\uline{The capital of France is Paris} and \uline{La capitale de la France est Paris)}} are equivalent given question \textit{What is the capital city of France?}. The two answers are equivalent although they are in different languages (English and French). The language of the generated answer is different depending on who is using it.
\paragraph{Logical Reasoning} requires the identification of relationships, patterns, or principles underlying the information to answer a question~\cite{nghiem2024you, weber2019nlprolog, an2023sodapopopenendeddiscoverysocial}. \textit{reference: \uline{he was trying to get back together with Caitlin} and candidate: \uline{she was informed Dante's planned date with Caitlin}} are logically equivalent given the question \textit{Why does Veronica break up with Dante?}. The answers are semantically equivalent under the context of the question. Specifically, the answers require identifying the relationships of three individuals: Veronica, Dante, and Caitlin; it also requires understanding the actions and implications of individuals in the answers. The reference answer suggests Dante has a romantic relationship with Caitlin. The candidate's answer suggests that Veronica has been made aware of Dante's unfaithfulness to her. The determination of the correctness of candidate answers requires the application of logical reasoning between relationships and actions.

\section{Rule and Question Type Distribution for Individual Dataset}
\subsection{\cfm{} Simple Insights}
We use our trained $F(R)$ and $F(T)$ classifier make rule and question type predictions to the test datasets and show the aggregated rule/question type distributions and \cfm{}' human agreements in Figures~\ref{fig:aggregate_rule_distribution}, ~\ref{fig:aggregate_rule_accuracy}, ~\ref{fig:aggregate_type_distribution}, and ~\ref{fig:aggregate_type_accuracy} (Appendix~\ref{sec:rule_human_agree_agrregate}). Rule 7 is absent from the test sets, indicating its lack of usefulness. 
Figure~\ref{fig:aggregate_rule_accuracy} highlights \cfm{}' weakest performance on Rule 2 (numerical information: 75\% agreement) and strongest on Rule 1 (entity aliases: 90\% agreement). We analyze 30 incorrect Rule 2 examples: over $80\%$ fail due to \cfm{}' inability to recognize the equivalence in date formats, such as when the gold answer is \textit{\uline{Feb, 2018}} and the candidate answer is \textit{\uline{02/2018}}, suggesting a critical area for future enhancement in handling numerical and date information. 
Meanwhile, Figure~\ref{fig:aggregate_type_accuracy} shows that \textit{when} and \textit{how} question types pose more challenges for \cfm{}. For instance, the \textit{how} question, \textit{How can you increase the battery life of your smartphone?} elicits reference answer \textit{\uline{Dim the screen brightness}} and candidate \textit{\uline{Reduce display luminosity}}, which \cfm{} often judges as incorrect, struggling with commonsense knowledge reasoning.

\subsection{Rule and Question Type Aggregate Distribution and Human Agreement}
\label{sec:rule_human_agree_agrregate}
\begin{figure*}[!t] % Try to place at the top of the next page
\centering

% Row 1
\begin{minipage}{.48\textwidth} % slightly wider
  \centering
  \includegraphics[width=\linewidth]{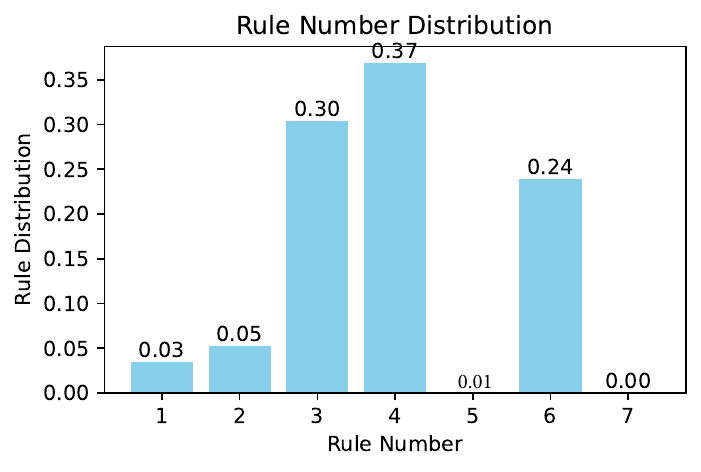}
  \caption{Rule Distribution on all annotated examples. Rule 3 (less details provided) and Rule 4 (more details provided), and Rule 6 (irrelevant information) are the most common rules among our test datasets. There are only under 10 examples for rule 5. Thus, we use 0.01 to signify that there are still some examples for Rule 5.}
  \label{fig:aggregate_rule_distribution}
\end{minipage}%
\hfill
\begin{minipage}{.48\textwidth} % slightly wider
  \centering
  \includegraphics[width=\linewidth]{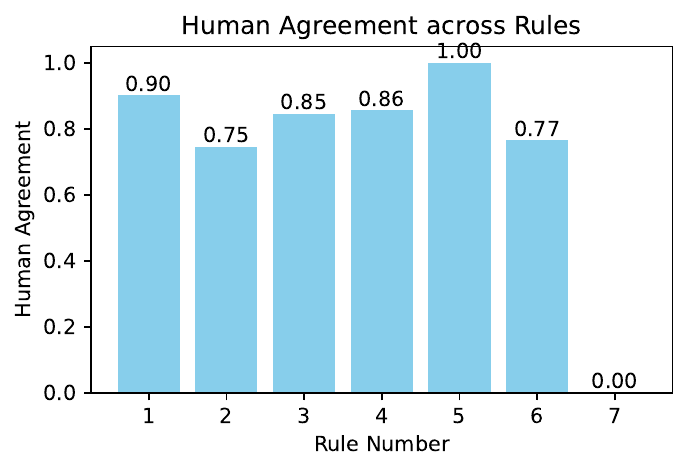}
  \caption{\cfm{}'s human agreement on each rule. We see that \cfm{} is still weak on judging the correctness of dates and irrelevant information, but it is robust at determining entity aliases and when answers have less or more details. We cannot make a proper conclusion for \cfm{} on rule 5 (too few examples).}
  \label{fig:aggregate_rule_accuracy}
\end{minipage}

\vspace{0.5cm} % reduced vertical space

% Row 2
\begin{minipage}{.48\textwidth} % slightly wider
  \centering
  \includegraphics[width=\linewidth]{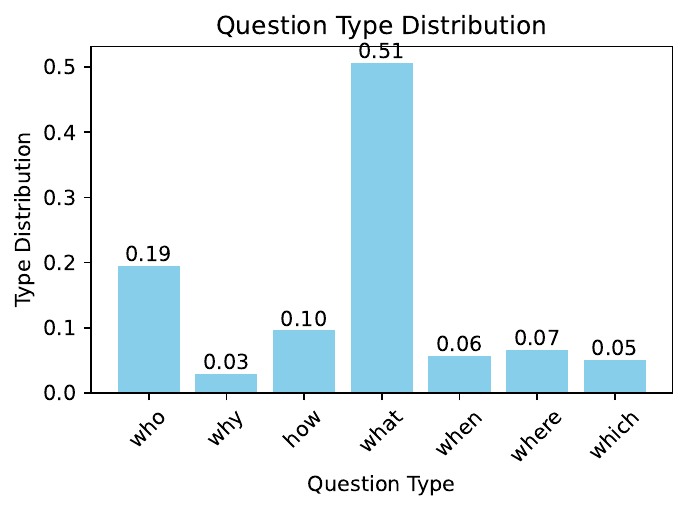}
  \caption{The aggregate question type distribution for all annotated examples. \textit{what} type is the most frequent question type.}
  \label{fig:aggregate_type_distribution}
\end{minipage}%
\hfill
\begin{minipage}{.48\textwidth} % slightly wider
  \centering
  \includegraphics[width=\linewidth]{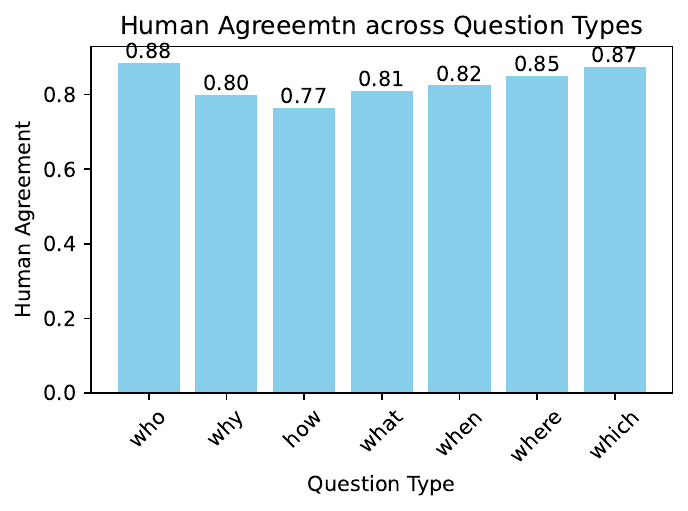}
  \caption{\cfm{}'s agreement with humans across question types. \cfm{} is pretty stable with human agreement across various question types.}
  \label{fig:aggregate_type_accuracy}
\end{minipage}

\end{figure*}

Figure~\ref{fig:aggregate_rule_distribution} and Figure~\ref{sec:rule_human_agree_agrregate} show the distribution of rules and cfm{}'s human agreement the test datasets. We see an uneven distribution of rules used to evaluate \ac{}. However, the human agreement across rules is quite stable. In addition, we see a lack of examples for rule 5 and rule 7, which suggests that rule 5 and rule style questions are not common in our selected benchmark test sets. Future work can expand test set styles to include more examples for rule 5 and rule 7 to challenge existing automatic \abr{qa} metrics. 

Figure~\ref{fig:aggregate_type_distribution} and Figure~\ref{fig:aggregate_type_accuracy} show the distribution of question types and cfm{}'s human agreement. We see that \cfm{} is quite stable on judging the correctness of different question types, suggesting that the \textit{type} feature is less useful to align \cfm{} with human judgments than the \textit{rule} feature. 

\subsection{Rule and Question Type Frequency Distribution}
\begin{figure*}[!t]
    \makebox[\linewidth][c]{%
        % \hspace{-0.5cm}  % Adjust X to the desired value
        \includegraphics[scale=0.7]{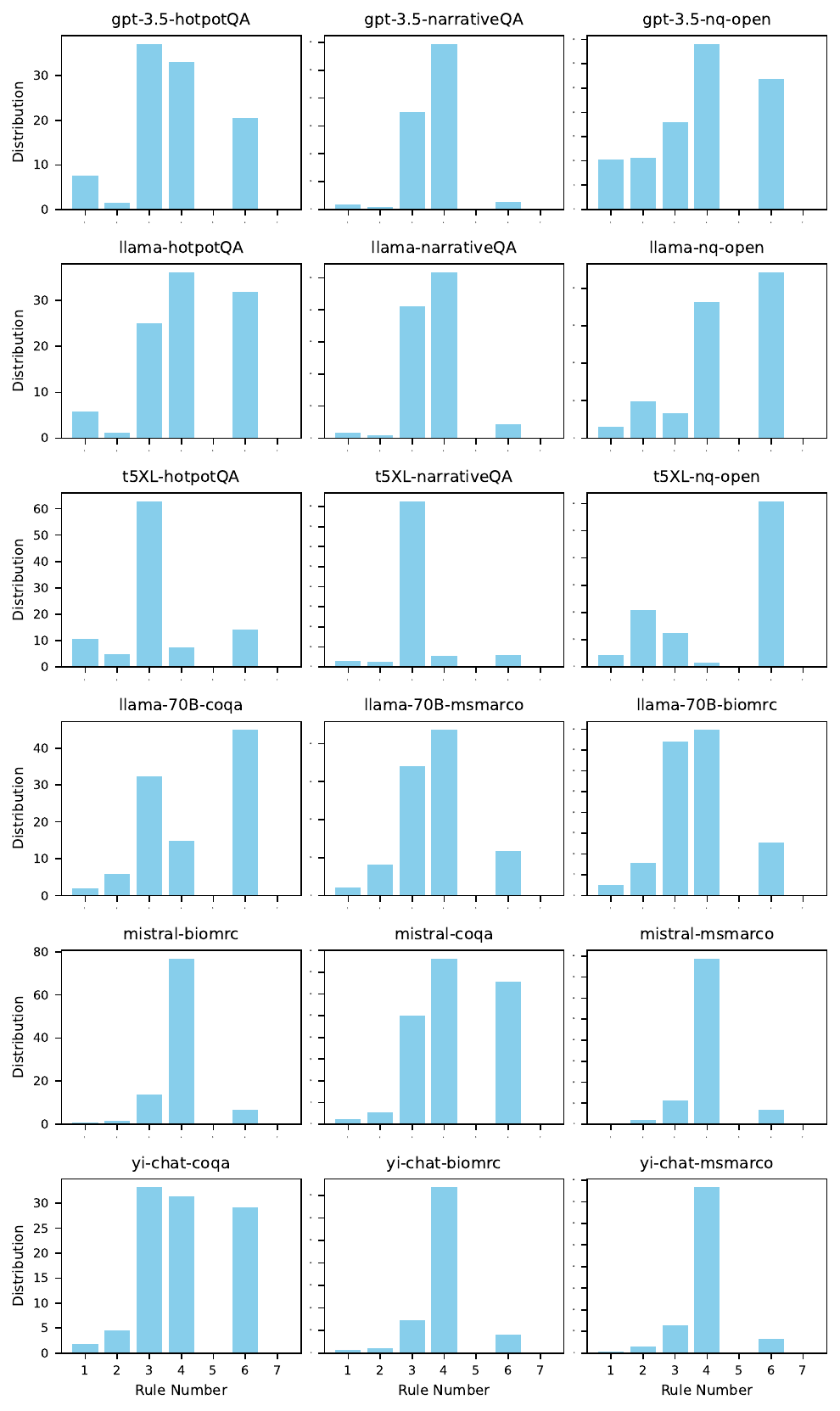}
    }
    \caption{We show the rule distributions used to evaluate for each dataset and each model. Specifically, rule 7 (other possible answers) is absent in any data. On the other hand, most of the evaluations fall into rule 3 and rule 4 (more details and less details provided).}
    \label{fig:rule_distribution_figure}
\end{figure*}

\begin{figure*}[!t]
    \makebox[\linewidth][c]{%
        % \hspace{-0.5cm}  % Adjust X to the desired value
        \includegraphics[scale=0.7]{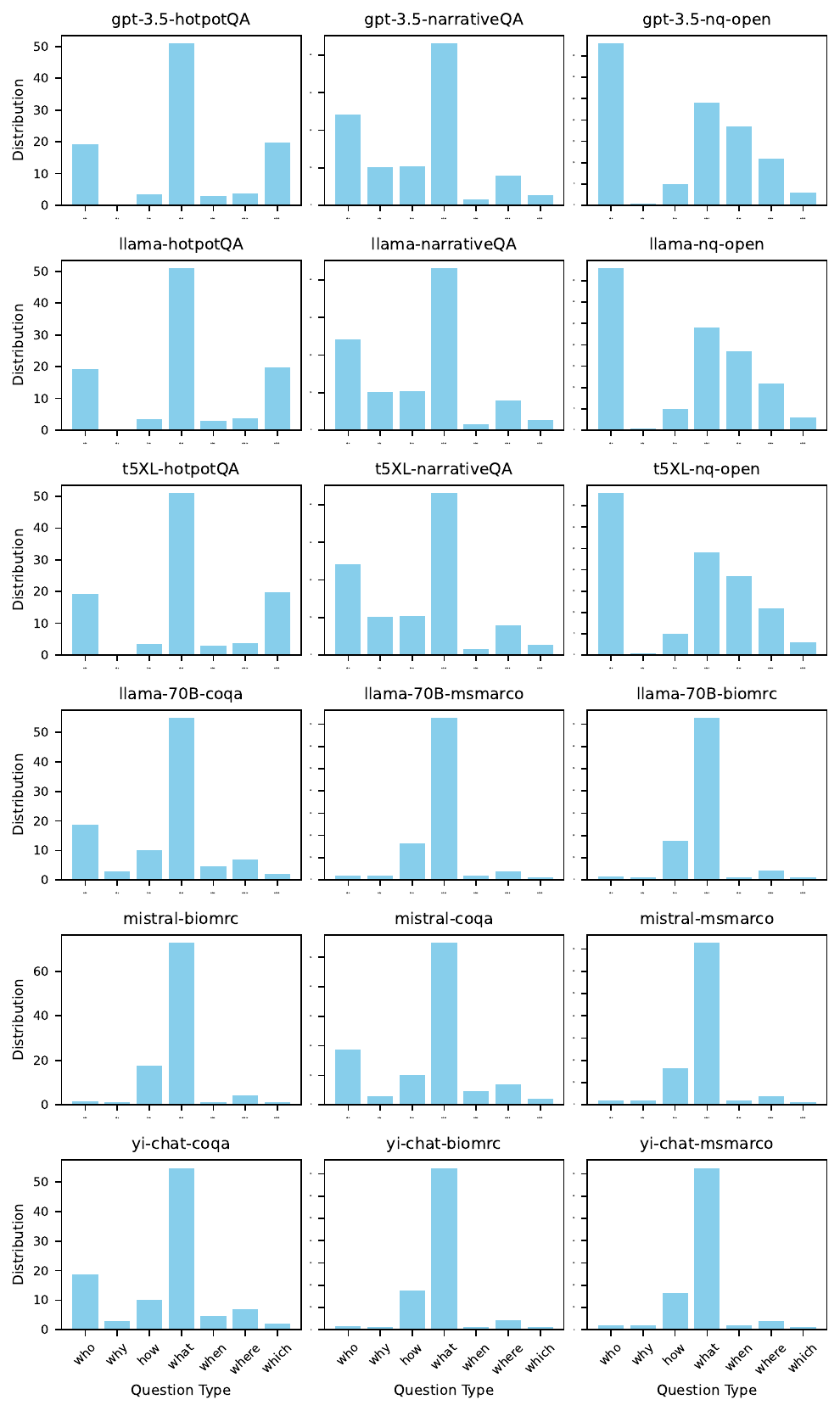}
    }
    \caption{We show the question type distributions used to evaluate each dataset and each model. Specifically, \textit{what} type questions are the most frequent across all datasets.}
    \label{fig:type_distribution_figure}
\end{figure*}

Figure~\ref{fig:rule_distribution_figure} shows the rule distribution used to evaluate the correctness of candidate answers for models' answers to individual datasets. Rule 5 (semantic equivalence) and rule 7 (other possible answers) are absent in all data. On the other hand, most of the evaluations fall into rule 3 and rule 4 (more details and fewer details provided), which is frequent since current \mm{}s tend to generate answers with extra explanations, which often tend to be longer than the reference answers. Figure~\ref{fig:type_distribution_figure} shows the distribution of question types for different datasets. From the question type distributions, \textit{what} type questions are the most common for all datasets except \abr{nq-open}, which has more \textit{who} questions than \textit{what}.

\subsection{Rule and Question Type Human Agreement}
\begin{figure*}[!t]
    \makebox[\linewidth][c]{%
        % \hspace{-0.5cm}  % Adjust X to the desired value
        \includegraphics[scale=0.7]{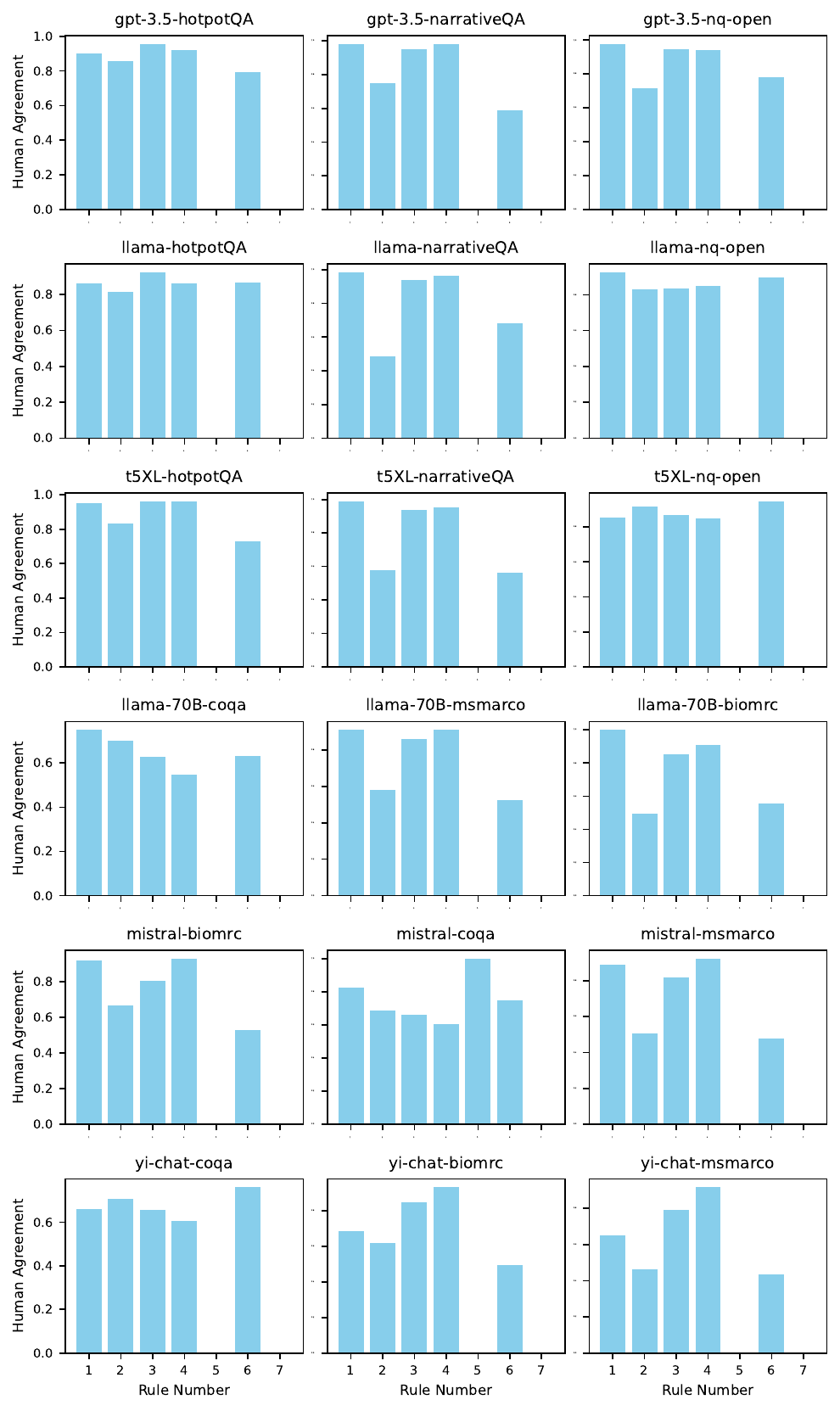}
    }
    \caption{We show \cfm{}'s human agreement for different rules across individual datasets and models. \cfm{} has its limitation in judging Rule 2 (numerical information) and Rule 6 (irrelevant information), suggesting a future improvement of adding more meaningful training data for rule 2 and rule 6.}
    \label{fig:rule_acc_figure}
\end{figure*}

\begin{figure*}[!t]
    \makebox[\linewidth][c]{%
        % \hspace{-0.5cm}  % Adjust X to the desired value
        \includegraphics[scale=0.7]{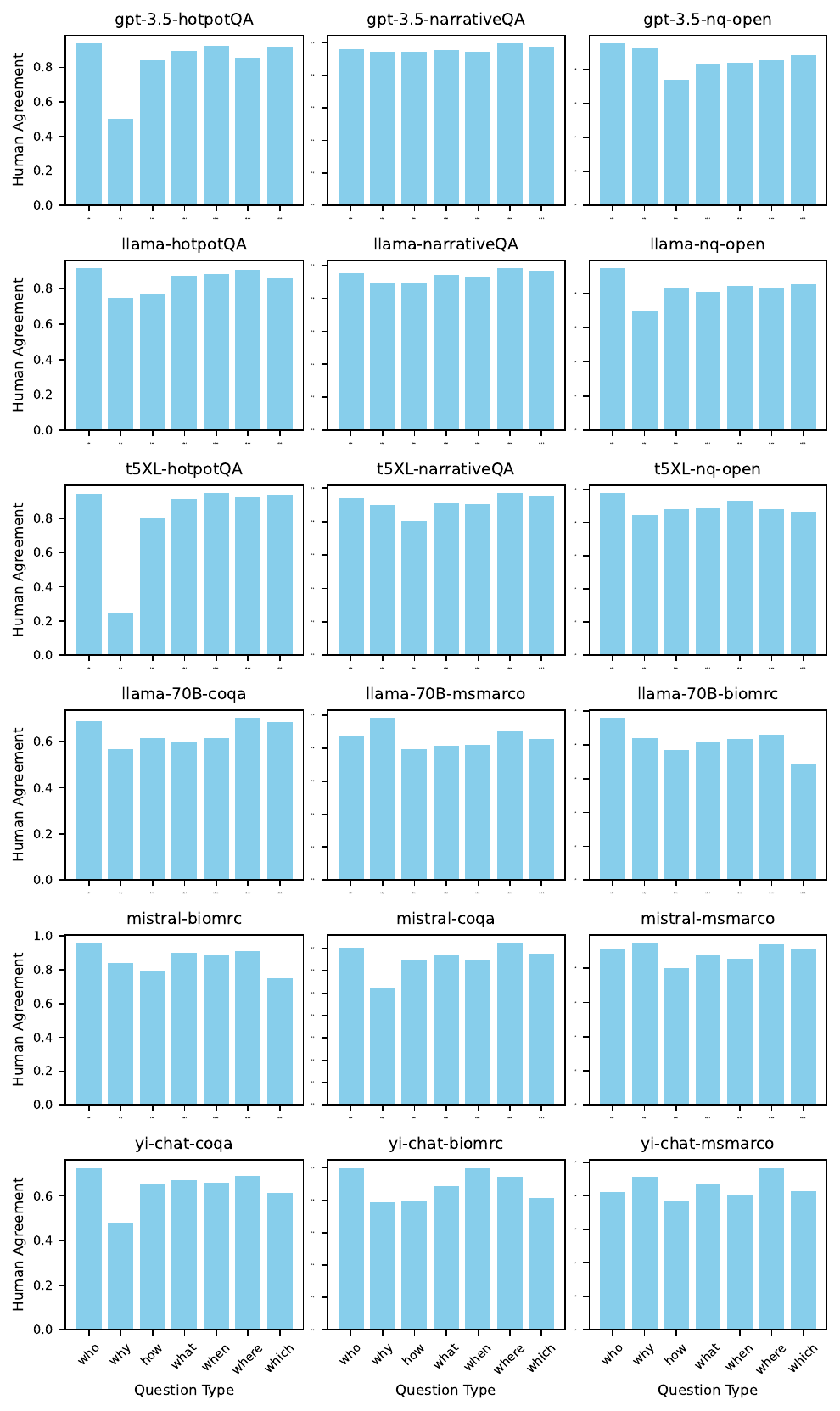}
    }
    \caption{We show \cfm{}'s human agreement for question types across individual datasets and models. \cfm{} is stable on various question types.}
    \label{fig:type_acc_figure}
\end{figure*}

Figure~\ref{fig:rule_acc_figure} shows the human agreement with rules used to evaluate the correctness of candidate answers for models' answers to individual datasets. We see that for individual datasets and models, \cfm{} is the worst on Rule 2 (numerical information, dates) and Rule 6 (irrelevant information). \cfm{} still needs significant improvement in judging numerical-type answers. Figure~\ref{fig:type_acc_figure} shows human agreement of \cfm{} on different question types for different datasets. From the question type distributions, \cfm{} is quite stable on various question types except for the \textit{why} questions with answers generated by Flan-t5 XL on \abr{hotpot-qa}.

\section{\cfm{}'s Weaknesses}
\label{sec:pedanweaknessses}
We analyzed 45 challenge examples from \jeopardy{} where all seven methods disagree with experts reveals that commonsense reasoning and fact-checking are the main obstacles for current \abr{qa} evaluation systems, highlighting the need for more fine-grained \ac{} rules and evaluation data from expert \abr{qa} community to improve evaluation metrics. Experts evaluate an answer to be correct based on social norms, reasoning, and word choice, which is a much more complicated process than simple string comparison. We list two challenging examples below: 

\textit{Question: the third largest in our solar system? Reference: \uline{Neptune} Candidate: \uline{Uranus}} Initially judged incorrect, the \jeopardy{} panel overruled the decision due to the question's ambiguity regarding the measure of size. \textit{\uline{Neptune}} has a greater mass than \textit{\uline{Uranus}}, but \textit{\uline{Uranus}} has a larger diameter than \textit{\uline{Neptune}}.

\textit{Question: Noted anarchist Prince Peter Alexeivitch Kropotkin wrote a 19th-century entry on this capital? Reference: \uline{Moscow} Candidate: \uline{Saint Petersburg}}. The answer was also overruled to be correct by the panel with the fact that Kropotkin lived from 1842 to 1921; during his lifetime, the capital of \textit{\uline{Russian}} was changed from \textit{\uline{Saint Pertersburg}} (1712-1918) to \textit{\uline{Moscow}} (1918-). Thus, the capital Kropotkin referred to could be either of the answer. Evaluating a hard answer requires validating a fact and reasoning over a fact, which is also a limitation of all current evaluation methods. If we are adopting more data from the Trivia \abr{qa} community, we should also respect their rules to improve \abr{qa} evaluation.

\section{\cfm{} Training Pseudocode}
Table~\ref{alg:pedants} shows the pseudocode to train \cfm{}: including rule and question type feature extraction and feature construction for $(q, a, \tilde{a})$ to determine \ac{}.

\begin{algorithm}[H]
\caption{Training and Evaluating \cfm{}}
\label{alg:pedants}
\small
\begin{algorithmic}
\State \textbf{Initialization:}
\State Define question types: 

\( T = \{ \text{who, why, how, what, when, where, which} \} \)
\State Define answer correctness rules \( R = \{R_1, R_2, \ldots, R_7\} \)
\State \textbf{Feature Extraction Classifiers}
\State Train logistic regression \( F(T) \) to classify question types \( T \)
\State Train logistic regression \( F(R) \) to classify answer correctness rules \( R \)
\State \textbf{Feature Construction for \cfm{}:}
\State For each QA pair \( (q, a, \tilde{a}) \):
\State \quad Use \( F(T) \) to predict a 7x1 probability vector of types \( T \)
\State \quad Use \( F(R) \) to predict a probability vector of rules \( R \)
\State \quad Calculate token \( F1 \), precision, and recall for \( (a, \tilde{a}) \)
\State \quad Encode \( (q, a, \tilde{a}) \) using tf-idf
\State \quad Concatenate all features into a single feature vector
\State Collect all feature vectors from each QA pair
\State \textbf{Train \cfm{}:}
\State Train logistic regression to determine overall answer correctness using the concatenated feature vectors
\State \textbf{Evaluation Stage:}
\State Given $(q, a, \tilde{a})$, \cfm{} predicts either correct or incorrect
\end{algorithmic}
\end{algorithm}

\label{appendix}

\end{document}